\documentclass[journal]{IEEEtran}
\usepackage{cite}
\usepackage{amsmath,amssymb,amsfonts,amsthm,amsxtra}
\usepackage{stmaryrd}
\usepackage[mathscr]{eucal}
\usepackage{commath}
\usepackage{algorithmic}
\usepackage{graphicx}
\usepackage{textcomp}
\usepackage{bm}
\usepackage[hidelinks=true]{hyperref}
\graphicspath{{images/}}
 
\newcommand{\argmin}{\operatornamewithlimits{argmin}}
\usepackage[ruled,linesnumbered,lined,commentsnumbered]{algorithm2e}
\usepackage{url}
\usepackage{array}

\usepackage{tasks}
\settasks{
	counter-format=(tsk[r]),
	label-width=3ex
}

\ifCLASSINFOpdf
\else
\fi
\begin{document}
%
\title{Sparse Nonnegative CANDECOMP/PARAFAC Decomposition in Block Coordinate Descent Framework: A Comparison Study}
%
%
%

\author{Deqing~Wang,~\IEEEmembership{Student~Member,~IEEE,}
        Fengyu~Cong,~\IEEEmembership{Senior~Member,~IEEE,}
        and~Tapani~Ristaniemi,~\IEEEmembership{Senior~Member,~IEEE}
\thanks{This work was supported in part by the National Natural Science Foundation of China (Grant No. 81471742), in part by the Fundamental Research Funds for the Central Universities [DUT16JJ(G)03] in Dalian University of Technology in China, and in part by the scholarship from China Scholarship Council (No. 201600090043). \textit{(Corresponding author: Fengyu Cong)}}
\thanks{D. Wang and F. Cong are with the School of Biomedical Engineering, Faculty of Electronic Information and Electrical Engineering, Dalian University of Technology, Dalian 116024, China, and also with the Faculty of Information Technology, University of Jyv\"askyl\"a, Jyv\"askyl\"a 40100, Finland (e-mail: deqing.wang@foxmail.com; cong@dlut.edu.cn).}
\thanks{T. Ristaniemi is with the Faculty of Information Technology, University of Jyv\"askyl\"a, Jyv\"askyl\"a 40100, Finland (e-mail: tapani.e.ristaniemi@jyu.fi).}
\thanks{Manuscript received Month Day, Year; revised Month Day, Year.}}

%
%

\markboth{Journal of \LaTeX\ Class Files,~Vol.~XX, No.~X, Month~2018}%
{Shell \MakeLowercase{\textit{et al.}}: Bare Demo of IEEEtran.cls for IEEE Journals}
%



\maketitle

\begin{abstract}
Nonnegative CANDECOMP/PARAFAC (NCP) decomposition is an important tool to process nonnegative tensor. Sometimes, additional sparse regularization is needed to extract meaningful nonnegative and sparse components. Thus, an optimization method for NCP that can impose sparsity efficiently is required. In this paper, we construct NCP with sparse regularization (sparse NCP) by $l_1$-norm. Several popular optimization methods in block coordinate descent framework are employed to solve the sparse NCP, all of which are deeply analyzed with mathematical solutions. We compare these methods by experiments on synthetic and real tensor data, both of which contain third-order and fourth-order cases. After comparison, the methods that have fast computation and high effectiveness to impose sparsity will be concluded. In addition, we proposed an accelerated method to compute the objective function and relative error of sparse NCP, which has significantly improved the computation of tensor decomposition especially for higher-order tensor.
\end{abstract}


%
\IEEEpeerreviewmaketitle

\section{Introduction}
%
%
%
%
\IEEEPARstart{N}{onnegative} tensor decomposition and nonnegative matrix factorization are powerful tools in signal processing and machine learning \cite{Cichocki2009,Wang2013,Zhou2014}. Due to the nonsubtractive property and part-based representation \cite{Lee1999}, they have been widely applied to hyperspectral unmixing \cite{He2017,Veganzones2016}, cognitive neuroscience \cite{MorupReport2006,Cong2015}, chemometrics \cite{Elcoroaristizabal2015,Vu2017}, and many other areas \cite{Wang2013,Morup2011}. Nonnegative CANDECOMP/PARAFAC (NCP) tensor decomposition is an extension of nonnegative matrix factorization (NMF) from two-way to multi-way. NMF and NCP are data factorization/decomposition methods with nonnegative constraints, which are described as follows.

\textbf{The NMF problem.} Given a nonnegative matrix $V\in\mathbb{R}^{I_{\bm{W}} \times I_{\bm{H}}}$ and a positive integer $R<\min \left\lbrace I_{\bm{W}}, I_{\bm{H}} \right\rbrace$, NMF is to solve the following minimization problem:
\begin{equation}
\begin{aligned}
\min_{\bm{W},\bm{H}}\frac{1}{2} & \norm{\bm{V}-\bm{W}\bm{H}}_F^2\\
\text{s.t. } \bm{W} & \geqslant 0, \bm{H} \geqslant 0,
\end{aligned}
\label{Eq:BasicNMF}
\end{equation}
where matrices $\bm{W}\in\mathbb{R}^{I_{\bm{W}} \times R}$ and $\bm{H}\in\mathbb{R}^{R \times I_{\bm{H}}}$ are two estimated nonnegative factors.

\textbf{The NCP problem.} Given a nonnegative $N$th-order tensor $\bm{\mathscr{X}}\in\mathbb{R}^{I_{1} \times I_{2} \times \dotsm \times I_{N}}$ and a positive number $R$, NCP is to solve the following minimization problem:
\begin{equation}
\begin{aligned}
\min_{\bm{A}^{(1)},\dotsc,\bm{A}^{(N)}} \frac{1}{2} & \norm{\bm{\mathscr{X}}-\llbracket \bm{A}^{(1)},\dotsc,\bm{A}^{(N)} \rrbracket}_{F}^{2}\\
\text{s.t. } \bm{A}^{(n)} & \geqslant 0 \text{ for } n=1, \dotsc,N,
\end{aligned}
\label{Eq:BasicNCP}
\end{equation}
where $\bm{A}^{(n)}\in\mathbb{R}^{I_{n} \times R}$ for $n=1,\dotsc,N$ are estimated factors on different modes, $I_{n}$ is the size on mode-$n$, and $R$ can be seen as the selected rank-1 tensor number (initial components number). The estimated factors in Kruskal operator can be represented by sum of $R$ rank-1 tensors in outer product form:
\begin{equation}
\llbracket \bm{A}^{(1)},\dotsc,\bm{A}^{(N)} \rrbracket = \sum_{r=1}^{R} \widetilde{\bm{\mathscr{X}}}_r = \sum_{r=1}^{R}\bm{a}_{r}^{(1)} \circ \dotsb \circ \bm{a}_{r}^{(N)}, 
\label{Eq:BasicNCP2}
\end{equation}
where $\bm{a}_{r}^{(n)}$ represents the $r$th column of $\bm{A}^{(n)}$.

Both NMF and NCP are non-convex and non-linear optimization problems, therefore finding their golobal minimums is NP-hard \cite{Kim2014}. Conventionally, problems \eqref{Eq:BasicNMF} and \eqref{Eq:BasicNCP} can be solved in block coordinate descent (BCD) framework \cite{Xu2013,Kim2014,Bertsekas2016}, in which each factor is updated alternatively as a subproblem with other factors fixed. These subproblems are usually convex. In NMF case, a lot of optimization methods have been proposed to solve these subproblems. Lee et al. proposed the multiplicative update (MU) method \cite{Lee1999,Lee2001}, which is the most popular and widely applied method for NMF. Berry et al. introduced the alternating least square (ALS) method in the review paper \cite{Berry2007}. Cichocki et al. proposed the hierarchical alternating least squares (HALS) method for large-scale NMF problems \cite{Cichocki2007,CichockiHALS2009}. Xu et al. proposed the alternating proximal gradient (APG) method for NMF with detailed mathematical convergence proofs \cite{Xu2013}. The same idea as APG was also proposed in \cite{Guan2012} for NMF independently, which was called NeNMF. Recently, the alternating direction method of multipliers (ADMM) has been employed for NMF problems \cite{Boyd2011,Sun2014}. In addition, the alternating nonnegative least squares (ANLS) method was deeply analyzed in Lin's seminal paper with strong optimization properties \cite{Lin2007}, which has a significant influence on NMF. When ANLS method is utilized, the subproblems of NMF appears as the nonnegative least squares (NNLS) problems. Many efficient methods have been devoted to solve the NNLS subproblems, such as Lin's project gradient method \cite{Lin2007}, quasi-Newton method \cite{Zdunek2006,Kim2007}, active-set method \cite{Kim2008}, block principal pivoting method \cite{Kim2011}, inertial projection neural network \cite{Dai2018}, and proximal function based method \cite{Zhang2014}. Most of above methods for NMF can be naturally extended to NCP problems \cite{Xu2013,Cichocki2009,CichockiHALS2009,Kim2012,Liavas2015,Huang2016}.

If no other constraint or regularization is imposed, \eqref{Eq:BasicNMF} and \eqref{Eq:BasicNCP} with only the nonnegative constraints can be seen as bound-constrained optimization problems \cite{Lin2007}. The nonnegative constraints can guarantee physically meaningful results. Sometimes, incorporating specific regularization can yield more interpretable and accurate results, e.g. sparse regularization, orthogonal regularization, smooth regularization \cite{Ito2016}, manifold regularization \cite{Cai2011,Shang2017}, and so on. Due to the intrinsic sparsity in many types of data, such as face image data \cite{Lee1999,Cai2011}, microarray data \cite{Kim2008}, text data \cite{Kim2011}, hyperspectral image data \cite{He2017} and EEG data \cite{Wang2018}, sparse results are always required in NMF and NCP problems. Nonnegativity constraint will naturally lead to sparsity in the results, but this sparsity in only a side effect, which can't be controlled to a certain level \cite{Hoyer2004}. Therefore, explicit sparse regularization is needed. In NMF case, many methods has been proposed to impose sparsity by projection or regularization iterms. Hoyer proposed to project all components in NMF factor into vectors that has desired sparsity degree \cite{Hoyer2004}, which can be solved by MU or projected gradient methods \cite{Mohammadiha2009}. However, this method makes all components have the same fixed sparsity degree, which doesn't reveal the true sparsity distribution in data. On the other hand, $l_1$-norm is a conventional and effective sparse regularizer for signal processing \cite{Bruckstein2009}. The reason is that for most underdetermined linear equations the minimization problem with $l_1$-norm regularization can yield strong sparsity \cite{Donoho2006}. Consequently, $l_1$-norm is a very promising regularization item for NMF and NCP to impose sparsity.

A multitude of works have been devoted to incorporate sparse regularization to NMF \cite{Hoyer2002,Kim2007,Kim2008,Xuan2018}, but rare works can be found for tensor decomposition. As far as we know, only a few studies had focused on imposing sparsity by $l_1$-norm regularization to Tucker decomposition \cite{Morup2008,Liu2012,Xu2015}. In this study, we investigate nonnegative CANDECOMP/PARAFAC (CP) decomposition with sparse regularization, which is abbreviated to ``sparse NCP" for convenience. We design the mathematical model of sparse NCP using $l_1$-norm as the engine to impose sparsity explicitly. In order to prevent rank deficiency and increase the stability, squared Frobenius norm is also added as an auxiliary regularization. The popular optimization methods of multiplicative update (MU), alternating least squares (ALS), hierarchical altering least squares (HALS), alternating proximal gradient (APG) and alternating nonnegative least squares (ANLS) are employed to solve the sparse NCP model, which are in block coordinate descent framework \cite{Xu2013,Kim2014,Bertsekas2016}. The sparse NCP implemented by all the above optimization methods are tested on synthetic tensor data and real tensor data, both of which contain third-order and fourth-order case. The abilities of these methods to impose sparsity are carefully compared. We want to mention HALS is a special method that includes normalization constraints on factor matrices in addition to nonnegative constraint. With additional normalization constraints, NMF and NCP no longer remain bounded problems \cite{Lin2007}. Despite the drawback of HALS, we still test its performance on the tensor data for imposing sparsity due to its popularity in the past \cite{Cichocki2009,CichockiHALS2009}.

Data in tensor, especially higher-order tensor (order $\geqslant$ 4), often consist of a huge amount of points, for which the decomposition might process slowly in some limited conditions. The iterating of tensor decomposition is usually terminated by checking the change of objective function value or data fitting, which is sometimes time consuming. A special strategy based on trace of matrix was used to accelerate the calculation of the objective function at each iteration \cite{Guan2012,Xu2013}. We extend this strategy to our sparse NCP problem, in which the $l_1$-norm sparse regularization and Frobenius regularization items can be easily handled.

The rest of this paper is organized as follows. In Section II, we describe the mathematical model of the designed sparse NCP. Section III elucidates the solutions to sparse NCP model using MU, ALS, HALS, APG, and ANLS method. In Section IV, we introduce our strategy to speed up the computation of sparse NCP. Section V describes the detailed experiments on synthetic and real datasets. Finally, we discuss several key issues related to sparse NCP in Section VI and conclude our paper in Section VI.

 
\section{Sparse Nonnegative CANDECOMP/PARAFAC Decomposition}
In this paper, operator $\circ$ represents outer product of vectors, $\odot$ represents the Khatri-Rao product,  $\ast$ represents the Hadamard product that is the elementwise matrix product, $\left\langle~\right\rangle$ represents inner product, and $\llbracket~\rrbracket$ represents Kruskal oprator. $\norm{~}_{F}$ denotes Frobenius norm, and $\norm{~}_1$ denotes $l_1$-norm. Basics of tensor computation and multi-linear algebra can be found in review \cite{Kolda2009}.

We present the designed sparse nonnegative CANDECOMP/PARAFAC decomposition (sparse NCP) in this section.

Given a nonnegative $N$th-order tensor $\bm{\mathscr{X}}\in\mathbb{R}^{I_{1} \times I_{2} \times \dotsm \times I_{N}}$ and a positive number $R$, we design the sparse NCP as the following minimization problem:
\begin{equation}
\begin{aligned}
\min_{\bm{A}^{(1)},\dotsc,\bm{A}^{(N)}} & \mathscr{O}= \frac{1}{2} \norm{\bm{\mathscr{X}}-\llbracket \bm{A}^{(1)},\dotsc,\bm{A}^{(N)} \rrbracket}_{F}^{2}\\
& +\sum_{n=1}^N\frac{\alpha_n}{2}\norm{\bm{A}^{(n)}}_F^2+\sum_{n=1}^N\beta_n\sum_{r=1}^R\norm{\bm{a}_r^{(n)}}_1\\
\text{s.t. } & \bm{A}^{(n)} \geqslant 0 \text{ for } n=1, \dotsc,N,
\end{aligned}
\label{Eq:SparseNCP}
\end{equation}
where $\bm{A}^{(n)}\in\mathbb{R}^{I_{n} \times R}$ for $n=1,\dotsc,N$ are the estimated factors in different modes, $\alpha_n$ and $\beta_n$ are positive regularization parameters in parameter vectors $\bm{\alpha}\in\mathbb{R}^{N\times 1}$ and $\bm{\beta}\in\mathbb{R}^{N\times 1}$, $I_{n}$ is the size in mode-$n$, $\bm{a}_{r}^{(n)}$ represents the $r$th column of $\bm{A}^{(n)}$, and $R$ is the initial number of components.

Let $\bm{X}_{(n)}\in \mathbb{R}^{I_n \times \prod_{\tilde{n}=1,\tilde{n}\neq n}^N I_{\tilde{n}}}$ represent the mode-$n$ unfolding (matricization) of original tensor $\bm{\mathscr{X}}$. And the mode-$n$ unfolding of the estimated tensor in Kruskal operator $\llbracket \bm{A}^{(1)},\dotsc,\bm{A}^{(N)} \rrbracket$  can be written as $\bm{A}^{(n)}{\left(\bm{B}^{(n)}\right)}^T$, in which $\bm{B}^{(n)}=\left(\bm{A}^{(N)}\odot\cdots\odot\bm{A}^{(n+1)}\odot\bm{A}^{(n-1)}\odot\cdots\odot\bm{A}^{(1)}\right)\in\mathbb{R}^{\prod_{\tilde{n}=1,\tilde{n}\neq n}^N I_{\tilde{n}} \times R}$. In block coordinate descent framework, factor $\bm{A}^{(n)}$ is updated alternatively by a subproblem in every iteration, which equals to the following minimization problem:
\begin{equation}
\begin{aligned}
\min_{\bm{A}^{(n)}} \mathscr{F} \left(\bm{A}^{(n)}\right) = & \frac{1}{2} \norm{\bm{X}_{(n)}-\bm{A}^{(n)}{\left(\bm{B}^{(n)}\right)}^T}_{F}^{2}\\
& +\frac{\alpha_n}{2}\norm{\bm{A}^{(n)}}_F^2+\beta_n\sum_{r=1}^R\norm{\bm{a}_r^{(n)}}_1\\
\text{s.t. } & \bm{A}^{(n)} \geqslant 0.
\end{aligned}
\label{Eq:SparseNCPSub}
\end{equation}
As previously mentioned, the sparse NCP problem in \eqref{Eq:SparseNCP} is non-convex, therefore finding its global minimum is NP-hard. However, the subproblem \eqref{Eq:SparseNCPSub} with Frobenius norm and $l_1$-norm is convex \cite{Guan2012}, which is the key point to solve \eqref{Eq:SparseNCP}. All the optimization methods introduced in the introduction can be applied to \eqref{Eq:SparseNCPSub}. Furthermore, the objective function in \eqref{Eq:SparseNCPSub} can be represented as
\begin{equation}
\begin{aligned}
\mathscr{F} \left(\bm{A}^{(n)}\right) & = \frac{1}{2} \text{tr} \left[ \bm{X}_{(n)}^T \bm{X}_{(n)} \right] - \text{tr} \left[{\left(\bm{A}^{(n)} \right)}^T \bm{X}_{n} \bm{B}^{(n)} \right]\\
& + \frac{1}{2} \text{tr} \left[   \bm{A}^{(n)}   {\left(\bm{B}^{(n)} \right)}^T    \bm{B}^{(n)}   {\left(\bm{A}^{(n)} \right)}^T    \right]\\
& + \frac{\alpha_n}{2} \text{tr} \left[ {\left(\bm{A}^{(n)} \right)}^T  \bm{A}^{(n)}  \right] + \beta_n \text{tr} \left[ \bm{E}^T \bm{A}^{(n)} \right],
\end{aligned}
\label{Eq:SparseNCPSubTr}
\end{equation}
where $\bm{E}\in\mathbb{R}^{I_n \times R}$ is a matrix whose elements are all equal to 1. The partial gradient (or partial derivative) of $\mathscr{F}$ with respect to $\bm{A}^{(n)}$ is always used during computation,
\begin{equation}
\begin{aligned}
& \nabla_{\bm{A}^{(n)}}\mathscr{F} \left(\bm{A}^{(n)}\right) = \frac{\partial}{\partial \bm{A}^{(n)}} \mathscr{F} \left(\bm{A}^{(n)}\right)\\
& = -\bm{X}_{(n)} \bm{B}^{(n)} + \bm{A}^{(n)} {\left( \bm{B}^{(n)} \right)}^T \bm{B}^{(n)} + \alpha_n \bm{A}^{(n)} + \beta_n \bm{E}\\
& =  \bm{A}^{(n)} \left[ {\left( \bm{B}^{(n)} \right)}^T \bm{B}^{(n)} + \alpha_n \bm{I}_R \right] -\bm{X}_{(n)} \bm{B}^{(n)} + \beta_n \bm{E},
\end{aligned}
\label{Eq:SparseNCPSubDeriv}
\end{equation}
where $\bm{X}_{(n)} \bm{B}^{(n)}$ is called the \textit{Matricized Tensor Times Khatri-Rao Product} (MTTKRP) \cite{Bader2008}. The item ${\left( \bm{B}^{(n)} \right)}^T \bm{B}^{(n)}$ can be computed efficiently by
\begin{equation}
\begin{aligned}
& {\left( \bm{B}^{(n)} \right)}^T \bm{B}^{(n)} = \left[ {\left( \bm{A}^{(N)} \right)}^T \bm{A}^{(N)} \right] \ast \cdots \\
& \ast \left[ {\left( \bm{A}^{(n+1)} \right)}^T \bm{A}^{(n+1)} \right] \ast \left[ {\left( \bm{A}^{(n-1)} \right)}^T \bm{A}^{(n-1)} \right]\\
& \ast \cdots \ast \left[ {\left( \bm{A}^{(1)} \right)}^T \bm{A}^{(1)} \right].
\end{aligned}
\label{Eq:BtB}
\end{equation}

\section{Optimization Methods for Solving Sparse NCP}
In this section, we present the solutions to the sparse NCP problem in \eqref{Eq:SparseNCP} by all the optimization methods of MU, ALS, HALS, APG, and ANLS. To the best of our knowledge, there are no existing solution proposed directly for \eqref{Eq:SparseNCP}. Some optimization methods had been employed to solve NMF and NCP with Frobenius norm or $l_1$-norm regularization item separately \cite{Berry2007,Ito2016,Kim2008,Kim2012}, so it is natural to combine the solution for these two items together.
\subsection{Multiplicative Update}
Multiplicative update (MU) was first proposed by Lee et al for NMF \cite{Lee1999,Lee2001}. Cai et al. proposed a straightforword way using lagrange multiplier to solve NMF subproblems \cite{Cai2011}, where the same update rules can be obtained as Lee's method. We extend Cai's method to tensor case. Represent tensor factor $\bm{A}^{(n)}$ elementwisely by $\bm{A}^{(n)}=\left[ a_{ir}^{(n)} \right]$ for $i=1,\dotsc,I_n$ and $r=1,\dotsc,R$. Let $\psi_{ir}$ be the Lagrange multiplier for constraint $a_{ir}^{(n)}\geqslant 0$, and $\bm{\Psi}=\left[\psi_{ir}\right]$, based on \eqref{Eq:SparseNCPSubTr} the Lagrange $\mathscr{L}$ is
\begin{equation}
\begin{aligned}
\mathscr{L} =~ & \mathscr{F} \left(\bm{A}^{(n)}\right)     +    \text{tr}\left[\bm{\Psi}{\left(\bm{A}^{(n)}\right)}^T\right]\\
=~ & \frac{1}{2} \text{tr} \left[ \bm{X}_{(n)}^T \bm{X}_{(n)} \right] - \text{tr} \left[{\left(\bm{A}^{(n)} \right)}^T \bm{X}_{n} \bm{B}^{(n)} \right]\\
& + \frac{1}{2} \text{tr} \left[   \bm{A}^{(n)}   {\left(\bm{B}^{(n)} \right)}^T    \bm{B}^{(n)}   {\left(\bm{A}^{(n)} \right)}^T    \right]\\
& + \frac{\alpha_n}{2} \text{tr} \left[ {\left(\bm{A}^{(n)} \right)}^T  \bm{A}^{(n)}  \right] + \beta_n \text{tr} \left[ \bm{E}^T \bm{A}^{(n)} \right]\\
& + \text{tr}\left[\bm{\Psi}{\left(\bm{A}^{(n)}\right)}^T\right].
\end{aligned}
\label{Eq:MU1}
\end{equation}
The partial derivative of $\mathscr{L}$ with respect to $\bm{A}^{(n)}$ is
\begin{equation}
\begin{aligned}
\frac{\partial \mathscr{L}}{\partial \bm{A}^{(n)}} = & -\bm{X}_{(n)} \bm{B}^{(n)} + \bm{A}^{(n)} {\left( \bm{B}^{(n)} \right)}^T \bm{B}^{(n)}\\
& + \alpha_n \bm{A}^{(n)} + \beta_n \bm{E} + \bm{\Psi}.
\end{aligned}
\label{Eq:MU2Deriv}
\end{equation}
Using KKT condition $\psi_{ir}a_{ir}^{(n)}=0$, we obtain the following equation for $a_{ir}^{(n)}$:
\begin{equation}
\begin{aligned}
& -{\left[ \bm{X}_{(n)} \bm{B}^{(n)} \right]}_{ir} a_{ir}^{(n)} + \\
& {\left[ \bm{A}^{(n)} {\left( \bm{B}^{(n)} \right)}^T \bm{B}^{(n)} + \alpha_n \bm{A}^{(n)} + \beta_n \bm{E} \right]}_{ir} a_{ir}^{(n)} = 0.
\end{aligned}
\label{Eq:MU3KKT}
\end{equation}
This equation leads to the following multiplicative updating rule:
\begin{equation}
\begin{aligned}
a_{ir}^{(n)} \leftarrow a_{ir}^{(n)} \frac{{\left[ \bm{X}_{(n)} \bm{B}^{(n)} \right]}_{ir}}{{\left[ \bm{A}^{(n)} \left(  {\left( \bm{B}^{(n)} \right)}^T \bm{B}^{(n)} + \alpha_n \bm{I}_R \right) + \beta_n \bm{E} \right]}_{ir}},
\end{aligned}
\label{Eq:MU4Update}
\end{equation}
where $\bm{I}_R\in\mathbb{R}^{R\times R}$ is a identity matrix.
The implementation of MU method is listed in \textbf{Algorithm \ref{Alg:MU}}.
\begin{algorithm}
\label{Alg:MU}
\caption{MU for sparse NCP in \eqref{Eq:SparseNCP}}
\SetKwInOut{Input}{Input}\SetKwInOut{Output}{Output}
\Input{$\bm{\mathscr{X}}$, $R$, $\bm{\alpha}$, $\bm{\beta}$}
\Output{$\bm{A}^{(n)}$, $n=1,\dotsc,N$}
Initialize $\bm{A}^{(n)}\in\mathbb{R}^{I_n \times R}$, $n=1,\dotsc,N$, using random numbers, then plus a small positive number $\epsilon$ to all elements\;
\Repeat{some termination criterion is reached}{
	\For{$n=1$ \KwTo $N$}{
		Make mode-$n$ unfolding of $\bm{\mathscr{X}}$ as $\bm{X}_{(n)}$\;
		Compute MTTKRP $\bm{X}_{(n)} \bm{B}^{(n)}$ and ${\big( \bm{B}^{(n)} \big)}^T \bm{B}^{(n)}$ based on \eqref{Eq:BtB}\;
		\For{$i=1$ \KwTo $I_n$}{
			\For{$r=1$ \KwTo $R$}{
			Update $a_{ir}^{(n)}$ according to \eqref{Eq:MU4Update}\;
			}
		}
	}
}
\end{algorithm}

\subsection{Alternating Least Squares}
Cichochi et al. comprehensively introduced the ALS method to solve NMF and NCP problems with diverse regularization items \cite{Cichocki2009}. ALS method is quite easy to implement for the sparse NCP problem \eqref{Eq:SparseNCP}.

Let the partial derivative of the objective function $\mathscr{F} \left(\bm{A}^{(n)}\right)$ in \eqref{Eq:SparseNCPSubDeriv} equal to 0,
\begin{equation}
\begin{aligned}
& \frac{\partial}{\partial \bm{A}^{(n)}} \mathscr{F} \left(\bm{A}^{(n)}\right)\\
& = \bm{A}^{(n)} \left[ {\left( \bm{B}^{(n)} \right)}^T \bm{B}^{(n)} + \alpha_n \bm{I}_R \right] - \left[ \bm{X}_{(n)} \bm{B}^{(n)} - \beta_n \bm{E} \right] \\
& = 0,
\end{aligned}
\label{Eq:ALS1Deriv}
\end{equation}
we have the following updating rule for $\bm{A}^{(n)}$,
\begin{equation}
\begin{aligned}
\bm{A}^{(n)} \leftarrow {\left[ \frac{\bm{X}_{(n)} \bm{B}^{(n)} - \beta_n \bm{E}}{{\left( \bm{B}^{(n)} \right)}^T \bm{B}^{(n)} + \alpha_n \bm{I}_R}  \right]}_+ ,
\end{aligned}
\label{Eq:ALS2Update}
\end{equation}
where ${\left[~\right]}_+$ is a half-wave rectifying nonlinear projection to enforce nonnegativity. \textbf{Algorithm \ref{Alg:ALS}} shows the implementation of ALS method.
\begin{algorithm}
\label{Alg:ALS}
\caption{ALS for sparse NCP in \eqref{Eq:SparseNCP}}
\SetKwInOut{Input}{Input}\SetKwInOut{Output}{Output}
\Input{$\bm{\mathscr{X}}$, $R$, $\bm{\alpha}$, $\bm{\beta}$}
\Output{$\bm{A}^{(n)}$, $n=1,\dotsc,N$}
Initialize $\bm{A}^{(n)}\in\mathbb{R}^{I_n \times R}$, $n=1,\dotsc,N$, using random numbers\;
\Repeat{some termination criterion is reached}{
	\For{$n=1$ \KwTo $N$}{
		Make mode-$n$ unfolding of $\bm{\mathscr{X}}$ as $\bm{X}_{(n)}$\;
		Compute MTTKRP $\bm{X}_{(n)} \bm{B}^{(n)}$ and ${\big( \bm{B}^{(n)} \big)}^T \bm{B}^{(n)}$ based on \eqref{Eq:BtB}\;
		Update $\bm{A}^{(n)}$ by $\bm{A}^{(n)} \leftarrow {\left[ \frac{\bm{X}_{(n)} \bm{B}^{(n)} - \beta_n \bm{E}}{{\left( \bm{B}^{(n)} \right)}^T \bm{B}^{(n)} + \alpha_n \bm{I}_R}  \right]}_+$\;
	}
}
\end{algorithm}

\subsection{Hierarchical Alternating Least Squares}
Hierarchical alternating least squares (HALS) is a method to update each factor column by column. For the sake of simplification, we use $\bm{a}_r$ and $\bm{b}_r$ instead of $\bm{a}_r^{(n)}$ and $\bm{b}_r^{(n)}$ in this part, which are the $r$th column of $\bm{A}^{(n)}$ and $\bm{B}^{(n)}$ respectively. We also use ${\left[ \bm{A}^{(n)} \right]}_{(:,r)}=\bm{a}_r\in\mathbb{R}^{I_n\times 1}$ to represent the column of a matrix, and ${\left[ \bm{A}^{(n)} \right]}_{(i,r)}=a_{ir}^{(n)}$ to represent an element in a matrix. The objective function $\mathscr{F}$ in \eqref{Eq:SparseNCPSub} can be represent as
\begin{equation}
\begin{aligned}
\mathscr{F}=\frac{1}{2}\norm{\bm{X}_{(n)}-\sum_{r=1}^R\bm{a}_r\bm{b}_r^T}_F^2+\frac{\alpha_n}{2}\sum_{r=1}^R\norm{\bm{a}_r}_2^2+\beta_n\norm{\bm{a}_r}_1.
\end{aligned}
\label{Eq:HALS1Obj}
\end{equation}
The minimization problem in \eqref{Eq:SparseNCPSub} can be solved iteratively by subproblems of columns:
\begin{equation}
\begin{aligned}
\min_{\bm{a}_r}\mathscr{F}_r=\frac{1}{2}\norm{\bm{Z}_r-\bm{a}_r\bm{b}_r^T}_F^2+ & \frac{\alpha_n}{2}\norm{\bm{a}_r}_2^2+\beta_n\norm{\bm{a}_r}_1\\
\text{s.t. }\bm{a}_r\geqslant 0&,
\end{aligned}
\label{Eq:HALS2SubObj}
\end{equation}
for $r=1,\dotsc,R$, and
\begin{equation}
\bm{Z}_r=\bm{X}_{(n)}-\sum_{\tilde{r}=1,\tilde{r}\neq r}^R\bm{a}_{\tilde{r}}\bm{b}_{\tilde{r}}^T.
\label{Eq:HALS3Residual1}
\end{equation}
The partial derivative of $\mathscr{F}_r$ with respect to $\bm{a}_r$ is
\begin{equation}
\begin{aligned}
\frac{\partial\mathscr{F}_r}{\partial\bm{a}_r} & =\left(\bm{a}_r\bm{b}_r^T-\bm{Z}_r\right)\bm{b}_r+\alpha_n\bm{a}_r+\beta_n\bm{1},\\
& = \left(\bm{b}_r^T\bm{b}_r+\alpha_n \right)\bm{a}_r-\left(\bm{Z}_r\bm{b}_r-\beta_n\bm{1} \right),
\end{aligned}
\label{Eq:HALS4Deriv}
\end{equation}
where $\bm{1}\in\mathbb{R}^{I_n\times 1}$ is a vector with all elements equaling to 1. When $\frac{\partial\mathscr{F}_r}{\partial\bm{a}_r}=0$, nonnagetive column vector $\bm{a}_r$ can be updated as
\begin{equation}
\begin{aligned}
\bm{a}_r\leftarrow\frac{{\left[ \bm{Z}_r\bm{b}_r-\beta_n\bm{1} \right]}_+}{\bm{b}_r^T\bm{b}_r+\alpha_n},
\end{aligned}
\label{Eq:HALS5Update1}
\end{equation}
which is a closed-form solution \cite{Kim2014}.

A fast HALS method was proposed to solve large-scale NMF problem \cite{Cichocki2009,Kim2014}. We use the same idea to solve the sparse NCP problem. $\bm{Z}_r$ in \eqref{Eq:HALS3Residual1} can also be represented as
\begin{equation}
\bm{Z}_r=\bm{X}_{(n)}-\sum_{\tilde{r}=1}^R\bm{a}_{\tilde{r}}\bm{b}_{\tilde{r}}^T+\bm{a}_r\bm{b}_r^T.
\label{Eq:HALS6Residual2}
\end{equation}
Replacing $\bm{Z}_r$ in \eqref{Eq:HALS5Update1} by \eqref{Eq:HALS6Residual2}, we obtain the new efficient update rule for $\bm{a}_r$ as is shown in \eqref{Eq:HALS7Update2}.
\newcounter{HALSUpdate}
\begin{figure*}[!t]
\normalsize
\setcounter{HALSUpdate}{\value{equation}}
\begin{equation}
\begin{aligned}
\bm{a}_r\leftarrow & \frac{{\left[ \left( \bm{X}_{(n)}-\sum_{\tilde{r}=1}^R\bm{a}_{\tilde{r}}\bm{b}_{\tilde{r}}^T+\bm{a}_r\bm{b}_r^T \right)   \bm{b}_r-\beta_n\bm{1} \right]}_+}{\bm{b}_r^T\bm{b}_r+\alpha_n}    =    \frac{{\left[ \bm{X}_{(n)}\bm{b}_r-\sum_{\tilde{r}=1}^R\bm{a}_{\tilde{r}}\bm{b}_{\tilde{r}}^T\bm{b}_r+\bm{a}_r\bm{b}_r^T\bm{b}_r-\beta_n\bm{1} \right]}_+}{\bm{b}_r^T\bm{b}_r+\alpha_n}\\
& =\frac{{\left[  {\left[\bm{X}_{(n)}\bm{B}^{(n)}\right]}_{(:,r)}   -    \bm{A}^{(n)}{\left[  {\left( \bm{B}^{(n)} \right)}^T \bm{B}^{(n)}\right]}_{(:,r)}   +   \bm{a}_r{\left[{\left( \bm{B}^{(n)} \right)}^T \bm{B}^{(n)}\right]}_{(r,r)}    -\beta_n\bm{1}\right]}_+}      {{\left[{\left( \bm{B}^{(n)} \right)}^T \bm{B}^{(n)}\right]}_{(r,r)}+\alpha_n}\\
& ={\left[    \frac{{\left[{\left( \bm{B}^{(n)} \right)}^T \bm{B}^{(n)}\right]}_{(r,r)}}{{\left[{\left( \bm{B}^{(n)} \right)}^T \bm{B}^{(n)}\right]}_{(r,r)}+\alpha_n}\bm{a}_r   +   \frac{{\left[\bm{X}_{(n)}\bm{B}^{(n)}\right]}_{(:,r)}   -    \bm{A}^{(n)}{\left[  {\left( \bm{B}^{(n)} \right)}^T \bm{B}^{(n)}\right]}_{(:,r)}    -\beta_n\bm{1}}{{\left[{\left( \bm{B}^{(n)} \right)}^T \bm{B}^{(n)}\right]}_{(r,r)}+\alpha_n}  \right]}_+
\end{aligned}
\label{Eq:HALS7Update2}
\end{equation}
\hrulefill
\vspace*{4pt}
\end{figure*}

However, as mentioned above, $\bm{a}_r$ in different factors is usually poorly scaled, therefore after updating in \eqref{Eq:HALS7Update2} it has to be normalized as
\begin{equation}
\begin{aligned}
\bm{a}_r^{(n)}\leftarrow\frac{\bm{a}_r^{(n)}}{\norm{\bm{a}_r^{(n)}}_2},
\end{aligned}
\label{Eq:HALS8Normalize}
\end{equation}
for $n=1,\dotsc,N-1$, and $r=1,\dotsc,R$. The procedures of HALS are illustrated in \textbf{Algorithm \ref{Alg:HALS}}.
\begin{algorithm}
\label{Alg:HALS}
\caption{HALS for sparse NCP in \eqref{Eq:SparseNCP}}
\SetKwInOut{Input}{Input}\SetKwInOut{Output}{Output}
\Input{$\bm{\mathscr{X}}$, $R$, $\bm{\alpha}$, $\bm{\beta}$}
\Output{$\bm{A}^{(n)}$, $n=1,\dotsc,N$}
Initialize $\bm{A}^{(n)}\in\mathbb{R}^{I_n \times R}$, $n=1,\dotsc,N$, using random numbers\;
\Repeat{some termination criterion is reached}{
	\For{$n=1$ \KwTo $N$}{
		Make mode-$n$ unfolding of $\bm{\mathscr{X}}$ as $\bm{X}_{(n)}$\;
		Compute MTTKRP $\bm{X}_{(n)} \bm{B}^{(n)}$ and ${\big( \bm{B}^{(n)} \big)}^T \bm{B}^{(n)}$ based on \eqref{Eq:BtB}\;
		\For{$r=1$ \KwTo $R$}{
			Update $\bm{a}_r^{(n)}$ using \eqref{Eq:HALS7Update2}\;
			\If{$n<N$}{
			Normalize $\bm{a}_r^{(n)}$ by $\bm{a}_r^{(n)}\leftarrow\frac{\bm{a}_r^{(n)}}{\norm{\bm{a}_r^{(n)}}_2}$\;
			}
		}
	}
}
\end{algorithm}

\subsection{Alternating Proximal Gradient}
The mathematical properties of alternating proximal gradient (APG) method were thoroughly analyzed in Xu's works \cite{Xu2013,Xu2015}. APG method has exhibited excellent performances on both NMF and NCP problems, which is also efficient to cope with $l_1$ sparse regularization \cite{Xu2015}. Supposing updating $\bm{A}^{(n)}$ in \eqref{Eq:SparseNCP} at the $k$th iteration, APG is computed as the following.

Calculate block-partial gradient of $\mathscr{F} \left(\bm{A}^{(n)}\right)$ in \eqref{Eq:SparseNCPSub} as $\nabla_{\bm{A}^{(n)}}\mathscr{F} \left(\bm{A}^{(n)}\right)=\frac{\partial}{\partial \bm{A}^{(n)}} \mathscr{F} \left(\bm{A}^{(n)}\right)$. We take
\begin{equation}
\begin{aligned}
L_{k-1}^{(n)}=\norm{{\left( \bm{B}_{k-1}^{(n)} \right)}^T \bm{B}_{k-1}^{(n)}}_2
\end{aligned}
\label{Eq:APG1Lipschetz1}
\end{equation}
as Lipschitz constant of $\nabla_{\bm{A}^{(n)}}\mathscr{F}$ with respect to $\bm{A}^{(n)}$, where $\norm{\bm{A}}_2$ is spectral norm of matrix. 
We also select
\begin{equation}
\begin{aligned}
\omega_{k-1}^{(n)}=\min \left( \hat{\omega}_{k-1},\delta_\omega\sqrt{\frac{L_{k-2}^{(n)}}{L_{k-1}^{(n)}}} \right),
\end{aligned}
\label{Eq:APG3Omega}
\end{equation}
where $\delta_\omega<1$ is predefined, and $\hat{\omega}_{k-1}=\frac{t_{k-1}-1}{t_k}$ with $t_0=1$, $t_k=\frac{1}{2}\left(1+\sqrt{1+4t_{k-1}^2}\right)$.

Let
\begin{equation}
\begin{aligned}
\bm{\widehat{A}}_{k-1}^{(n)}=\bm{A}_{k-1}^{(n)}+\omega_{k-1}^{(n)}\left(\bm{A}_{k-1}^{(n)}-\bm{A}_{k-2}^{(n)}  \right)
\end{aligned}
\label{Eq:APG4ExtrapPnt}
\end{equation}
denote an extrapolated point where $\omega_{k-1}^{(n)}$ is the extrapolation weight, and let
\begin{equation}
\begin{aligned}
\bm{\widehat{G}}_{k-1}^{(n)}= & \bm{\widehat{A}}_{k-1}^{(n)} \left[ {\left( \bm{B}_{k-1}^{(n)} \right)}^T \bm{B}_{k-1}^{(n)} + \alpha_n\bm{I}_R \right]\\
& - \bm{X}_{(n)} \bm{B}^{(n)}+ \beta_n \bm{E}
\end{aligned}
\label{Eq:APG5ParGradient}
\end{equation}
represent the block-partial gradient of $\mathscr{F}\left(\bm{A}^{(n)}\right)$ at $\bm{\widehat{A}}_{k-1}^{(n)}$. Factor $\bm{A}^{(n)}$ at iteration $k$ is updated by
\begin{equation}
\begin{aligned}
\bm{A}_k^{(n)}=\argmin_{\bm{A}^{(n)}\geqslant 0} & \left\langle\bm{\widehat{G}}_{k-1}^{(n)},\bm{A}^{(n)}-\bm{\widehat{A}}_{k-1}^{(n)} \right\rangle\\
& + \frac{L_{k-1}^{(n)}}{2} \norm{\bm{A}^{(n)}-\bm{\widehat{A}}_{k-1}^{(n)}}_F^2.
\end{aligned}
\label{Eq:APG6Update1}
\end{equation}
The closed form of \eqref{Eq:APG6Update1} can be written as
\begin{equation}
\begin{aligned}
\bm{A}_k^{(n)} =\max\Bigg(0, & ~ \bm{\widehat{A}}_{k-1}^{(n)}-\frac{\bm{\widehat{G}}_{k-1}^{(n)}}{L_{k-1}^{(n)}} \Bigg)\\
= \max \Bigg(       0,  \bm{\widehat{A}}_{k-1}^{(n)} & -  \frac{\bm{\widehat{A}}_{k-1}^{(n)} \left[ {\left( \bm{B}_{k-1}^{(n)} \right)}^T \bm{B}_{k-1}^{(n)} + \alpha_n\bm{I}_R \right]}{L_{k-1}^{(n)}}\\
&+  \frac{\bm{X}_{(n)} \bm{B}^{(n)} - \beta_n \bm{E}}{L_{k-1}^{(n)}} \Bigg).
\end{aligned}
\label{Eq:APG7Update2}
\end{equation}

APG method for sparse NCP can be implemented by procedures in \textbf{Algorithm \ref{Alg:APG}}.
\begin{algorithm}
\label{Alg:APG}
\caption{APG for sparse NCP in \eqref{Eq:SparseNCP}}
\SetKwInOut{Input}{Input}\SetKwInOut{Output}{Output}
\Input{$\bm{\mathscr{X}}$, $R$, $\bm{\alpha}$, $\bm{\beta}$, $\delta_\omega$}
\Output{$\bm{A}^{(n)}$, $n=1,\dotsc,N$}
Initialize $\bm{A}^{(n)}\in\mathbb{R}^{I_n \times R}$, $n=1,\dotsc,N$, using random numbers\;
\For{$k=1,2,\dotsc$}{
	\For{$n=1$ \KwTo $N$}{
		Make mode-$n$ unfolding of $\bm{\mathscr{X}}$ as $\bm{X}_{(n)}$\;
		Compute MTTKRP $\bm{X}_{(n)} \bm{B}^{(n)}$ and ${\big( \bm{B}^{(n)} \big)}^T \bm{B}^{(n)}$ based on \eqref{Eq:BtB}\;
		Compute $L_{k-1}^{(n)}$, $\omega_{k-1}^{(n)}$, $\bm{\widehat{A}}_{k-1}^{(n)}$, $\bm{\widehat{G}}_{k-1}^{(n)}$ according to \eqref{Eq:APG1Lipschetz1}, \eqref{Eq:APG3Omega}, \eqref{Eq:APG4ExtrapPnt}, \eqref{Eq:APG5ParGradient}\;
		Update $\bm{A}_k^{(n)}$ according to \eqref{Eq:APG7Update2}\;
	}
	\If{$\mathscr{O}\left(\bm{A}_k^{(n)}\right)>\mathscr{O}\left(\bm{A}_{k-1}^{(n)}\right)$}{
		$\bm{\widehat{A}}_{k-1}^{(n)}$=$\bm{A}_{k-1}^{(n)}$, for $n=1,\dotsc,N$\;
		Update $\bm{A}_k^{(n)}$ again according to \eqref{Eq:APG7Update2}\;
	}
	\If{some termination criterion is reached}{
		\Return $\bm{A}_k^{(n)}$, for $n=1,\dotsc,N$\;
	}
}
\end{algorithm}

\subsection{Alternating Nonnegative Least Squares}

Alternating Nonnegative Least Squares (ANLS) is an important and efficient method for NMF problems \cite{Lin2007,Kim2008,Kim2011}. Kim, H. and Kim, J. utilized ANLS to solve NMF with squared Frobenius-norm reqularization and squared $l_1$-norm sparse regularization \cite{Kim2008,Kim2011}. This idea can be naturally extended to the following sparsity regularized NCP problem:
\begin{equation}
\begin{aligned}
\min_{\bm{A}^{(1)},\dotsc,\bm{A}^{(N)}} & \mathscr{O}_{\text{ANLS}}= \frac{1}{2} \norm{\bm{\mathscr{X}}-\llbracket \bm{A}^{(1)},\dotsc,\bm{A}^{(N)} \rrbracket}_{F}^{2}\\
+\sum_{n=1}^N & \frac{\alpha_n}{2}\norm{\bm{A}^{(n)}}_F^2+\sum_{n=1}^N\frac{\beta_n}{2}\sum_{r=i}^{I_n}\norm{{\left[\bm{A}^{(n)} \right]}_{(i,:)} }_1^2\\
\text{s.t. } & \bm{A}^{(n)} \geqslant 0 \text{ for } n=1, \dotsc,N,
\end{aligned}
\label{Eq:ANLS1}
\end{equation}
where ${\left[\bm{A}^{(n)} \right]}_{(i,:)}$ denotes the $i$th row of $\bm{A}^{(n)}$. Factors $\bm{A}^{(n)}$ can be updated alternatively by the following minimization subproblem:
\begin{equation}
\begin{aligned}
\min_{\bm{A}^{(n)}} \mathscr{F}_{\text{NNLS}} \left(\bm{A}^{(n)}\right) = & \frac{1}{2} \norm{\bm{X}_{(n)}-\bm{A}^{(n)}{\left(\bm{B}^{(n)}\right)}^T}_{F}^{2}\\
+\frac{\alpha_n}{2}\norm{\bm{A}^{(n)}}_F^2 & +\frac{\beta_n}{2}\sum_{r=i}^{I_n}\norm{{\left[\bm{A}^{(n)} \right]}_{(i,:)} }_1^2\\
\text{s.t. } & \bm{A}^{(n)} \geqslant 0.
\end{aligned}
\label{Eq:ANLS2Sub1}
\end{equation}
By some simple mathematical operations, \eqref{Eq:ANLS2Sub1} is equivalent to the following minimization problem:
\begin{equation}
\begin{aligned}
\min_{\bm{A}^{(n)}} & \mathscr{F}_{\text{NNLS}} \left(\bm{A}^{(n)}\right) =\\
& \frac{1}{2}
\norm{
\begin{pmatrix}
\bm{B}^{(n)}\\
\sqrt{\alpha_n}\bm{I}_R\\
\sqrt{\beta_n}\bm{1}_{1\times R}
\end{pmatrix}
{\left(\bm{A}^{(n)}\right)}^T-
\begin{pmatrix}
\bm{X}_{(n)}^T\\
\bm{0}_{R\times I_n}\\
\bm{0}_{1\times I_n}
\end{pmatrix}
}_F^2.\\
& \text{s.t. } \bm{A}^{(n)} \geqslant 0,
\end{aligned}
\label{Eq:ANLS3Sub2}
\end{equation}
Afterwards, the partial derivative of $\mathscr{F}_{\text{NNLS}} \left(\bm{A}^{(n)}\right)$ to $\bm{A}^{(n)}$ is
\begin{equation}
\begin{aligned}
&  \frac{\partial}{\partial \bm{A}^{(n)}} \mathscr{F}_{\text{NNLS}} \left(\bm{A}^{(n)}\right)\\
& =  \bm{A}^{(n)} \left[ {\left( \bm{B}^{(n)} \right)}^T \bm{B}^{(n)} + \alpha_n \bm{I}_R + \beta_n \bm{E} \right] -\bm{X}_{(n)} \bm{B}^{(n)}.
\end{aligned}
\label{Eq:ANLS4SubDer}
\end{equation}
From \eqref{Eq:ANLS3Sub2} and \eqref{Eq:ANLS4SubDer}, it is clear to see that the subproblem in \eqref{Eq:ANLS2Sub1} still satisfies the basic nonnegative least squares (NNLS) structure, which can be solve conveniently by those optimization methods for NNLS, such as active-set (AS) \cite{Kim2008} and block principal pivoting (BPP) \cite{Kim2011}.

In \eqref{Eq:ANLS1}, the squared $l_1$-norm of the rows in $\bm{A}^{(n)}$ is used to impose sparsity in order to satisfy the NNLS structure, which is different from the sparse NCP in \eqref{Eq:SparseNCP}. \textbf{Algorithm \ref{Alg:ANLS}} explicates the ANLS method for sparsity regularized NCP in \eqref{Eq:ANLS1}.

\begin{algorithm}
\label{Alg:ANLS}
\caption{ANLS for sparsity regularized NCP in \eqref{Eq:ANLS1}}
\SetKwInOut{Input}{Input}\SetKwInOut{Output}{Output}
\Input{$\bm{\mathscr{X}}$, $R$, $\bm{\alpha}$, $\bm{\beta}$}
\Output{$\bm{A}^{(n)}$, $n=1,\dotsc,N$}
Initialize $\bm{A}^{(n)}\in\mathbb{R}^{I_n \times R}$, $n=1,\dotsc,N$, using random numbers\;
\Repeat{some termination criterion is reached}{
	\For{$n=1$ \KwTo $N$}{
		Make mode-$n$ unfolding of $\bm{\mathscr{X}}$ as $\bm{X}_{(n)}$\;
		Compute MTTKRP $\bm{X}_{(n)} \bm{B}^{(n)}$ and ${\big( \bm{B}^{(n)} \big)}^T \bm{B}^{(n)}$ based on \eqref{Eq:BtB}\;
		${\big( \bm{B}^{(n)} \big)}^T \bm{B}^{(n)} \leftarrow {\big( \bm{B}^{(n)} \big)}^T \bm{B}^{(n)} + \alpha_n \bm{I}_R + \beta_n \bm{E}$\;
		Update factor $\bm{A}^{(n)}$ using NNLS method based on \eqref{Eq:ANLS3Sub2}:\\

		$\displaystyle \bm{A}^{(n)} = \argmin_{\bm{A}^{(n)}\geqslant 0} \mathscr{F}_{\text{NNLS}} \left(\bm{A}^{(n)}\right)$\\
		$\displaystyle =\texttt{NNLS\_AS(} \bm{X}_{(n)} \bm{B}^{(n)} \texttt{,} {\big( \bm{B}^{(n)} \big)}^T \bm{B}^{(n)} \texttt{)}$\\
		or\\
		$\displaystyle =\texttt{NNLS\_BPP(} \bm{X}_{(n)} \bm{B}^{(n)} \texttt{,} {\big( \bm{B}^{(n)} \big)}^T \bm{B}^{(n)} \texttt{)}$.
	}
}
\end{algorithm}

\section{Stopping Condition and Accelerating Strategy}
The optimization procedures for tensor decomposition are implemented by iterations. For NCP, a sequence of ${\left\lbrace \bm{A}_k^{(1)}, \dotsc, \bm{A}_k^{(N)} \right\rbrace}_{k=1}^\infty$ is produced at each iteration. It is necessary to terminate the iteration until some stopping condition is satisfied. Common stopping conditions includes the following: predefined maximum number of iterations, predefined maximum running time, the change of objective function value, the change of relative error (data fitting) \cite{Kolda2009,Xu2013}.

\subsection{Objective Function and Relative Error}

The computations of objective function value and relative error (data fitting) are highly correlated. In the $k$th iteration, the objective function value of NCP problem \eqref{Eq:BasicNCP} is
\begin{equation}
\mathscr{O}_{\text{NCP}k} = \frac{1}{2} \norm{\bm{\mathscr{X}}-\llbracket \bm{A}_k^{(1)},\dotsc,\bm{A}_k^{(N)} \rrbracket}_{F}^{2},
\label{Eq:BasicNCPObj}
\end{equation}
and the relative error \cite{Xu2013} is defined by
\begin{equation}
\text{RelErr}_k=\frac{\norm{\bm{\mathscr{X}} - \llbracket \bm{A}_k^{(1)},\dotsc,\bm{A}_k^{(N)} \rrbracket}_F}{\norm{\bm{\mathscr{X}}}_F}.
\label{Eq:RelErr}
\end{equation}
Comparing \eqref{Eq:BasicNCPObj} and \eqref{Eq:RelErr}, the relative error can also be computed from the objective function value directly:
\begin{equation}
\text{RelErr}_k= \frac{\sqrt{2\mathscr{O}_{\text{NCP}k}}}{\norm{\bm{\mathscr{X}}}_F}.
\label{Eq:RelErr2}
\end{equation}
Meanwhile, the data fitting can be computed by
\begin{equation}
\text{Fit}_k=1-\text{RelErr}_k.
\label{Eq:DataFitting}
\end{equation}

Based on the objective function value and the relative error, the stopping condition can be set by
\begin{equation}
\lvert \text{RelErr}_{k-1}-\text{RelErr}_k \rvert < \epsilon
\label{Eq:StopRelErr}
\end{equation}
or
\begin{equation}
\lvert \mathscr{O}_{\text{NCP}k-1}-\mathscr{O}_{\text{NCP}k} \rvert < \epsilon.
\label{Eq:StopObj}
\end{equation}
The threshold of $\epsilon$ can be set by a very small positive value, such as $1e-8$.

\subsection{Accelerated Computation of Objective Function}
By mode-$n$ unfolding of tensor, the objective function of NCP in \eqref{Eq:BasicNCPObj} at the $k$th iteration can be represented equivalently as the following:
\begin{equation}
\mathscr{O}_{\text{NCP}k}= \frac{1}{2} \norm{\bm{X}_{(n)}-\bm{A}_k^{(n)}{\left(\bm{B}_k^{(n)}\right)}^T}_{F}^{2}.
\label{Eq:BasicNCPObjMatrix}
\end{equation}
The works of \cite{Guan2012} and \cite{Xu2013} introduced a convenient idea to compute the objective function base on the trace computation of matrix. Inspired by this idea, we further represent the objective function in \eqref{Eq:BasicNCPObjMatrix} by
\begin{equation}
\begin{aligned}
&\mathscr{O}_{\text{NCP}k}\\
&=\frac{1}{2} \text{tr} \Bigg\lbrace {\left\lbrack \bm{X}_{(n)} - \bm{A}_k^{(n)} {\left(\bm{B}_k^{(n)}\right)}^T \right\rbrack}^T {\left\lbrack \bm{X}_{(n)} - \bm{A}_k^{(n)} {\left(\bm{B}_k^{(n)}\right)}^T \right\rbrack} \Bigg\rbrace\\
&=\frac{1}{2} \Bigg\lbrace \norm{\bm{\mathscr{X}}}_F^2 - 2\text{tr} \left\lbrack \bm{A}_k^{(n)} {\left(\bm{X}_{(n)} \bm{B}_k^{(n)} \right)}^T \right\rbrack\\
&+\text{tr} \left\lbrack \left( {\left(\bm{A}_k^{(n)}\right)}^T \bm{A}_k^{(n)} \right) \left( {\left(\bm{B}_k^{(n)}\right)}^T \bm{B}_k^{(n)} \right) \right\rbrack \Bigg\rbrace.
\end{aligned}
\label{Eq:BasicNCPObjTrace}
\end{equation}
Furthermore, the objective function equals to
\begin{equation}
\mathscr{O}_{\text{NCP}k}=\frac{1}{2} \Bigg\lbrace \norm{\bm{\mathscr{X}}}_F^2 - 2\sum_{j=1}^R\sum_{i=1}^{I_n} \widehat{\bm{N}}_{i,j} + \sum_{j=1}^R \sum_{i=1}^R \widehat{\bm{M}}_{i,j} \Bigg\rbrace,
\label{Eq:BasicNCPObjFast}
\end{equation}
where $\widehat{\bm{N}}=\bm{A}_k^{(n)} \ast {\left(\bm{X}_{(n)} \bm{B}_k^{(n)} \right)} \in \mathbb{R}^{I_n \times R}$ and $\widehat{\bm{M}}=\left( {\left(\bm{A}_k^{(n)}\right)}^T \bm{A}_k^{(n)} \right) \ast \left( {\left(\bm{B}_k^{(n)}\right)}^T \bm{B}_k^{(n)} \right) \in \mathbb{R}^{R \times R}$. Here, $\ast$ is the Hadamard product.

Since tensor data usually consist of a large amount of data points, the computation of the objective function at each iteration will be time consuming by \eqref{Eq:BasicNCPObj} or \eqref{Eq:BasicNCPObjMatrix}. For example, in \eqref{Eq:BasicNCPObjMatrix} the computational complexity of $\bm{A}_k^{(n)}{\left(\bm{B}_k^{(n)}\right)}^T$ is $O(R \times \prod_{n=1}^N I_n)$. On the other hand, by observing \textbf{Algorithm \ref{Alg:MU}} to \textbf{\ref{Alg:ANLS}}, we find that the items of $\bm{X}_{(n)} \bm{B}_k^{(n)}$ and ${\left(\bm{B}_k^{(n)}\right)}^T \bm{B}_k^{(n)}$ have been computed in advance in order to update $\bm{A}_k^{(n)}$. Therefore, these two items can be employed directly to compute the objective function by \eqref{Eq:BasicNCPObjFast}. In \eqref{Eq:BasicNCPObjFast}, the computational complexity of $\widehat{\bm{N}}$ and $\widehat{\bm{M}}$ is only $O(I_nR+R^2)$, which has been reduced significantly.

\subsection{Computation of Objective Function for Sparse NCP}
The objective function and relative error for sparse NCP should be carefully considered, due to the extra Frobenius norm and $l_1$-norm regularization items. Based on the relationship introduced in Subsection IV-A and the accelerating method in Subsection IV-B, the computation of objective function and relative error for sparse NCP is explicated in \textbf{Algorithm \ref{Alg:ObjRelErr}}.

\begin{algorithm}
\label{Alg:ObjRelErr}
\caption{Objective function and relative error for sparse NCP}
\SetKwInOut{Input}{Input}\SetKwInOut{Output}{Output}
\Input{$\bm{\mathscr{X}}$, $R$, $\bm{\alpha}$, $\bm{\beta}$, $\epsilon$}
\Output{$\bm{A}^{(n)}$, $n=1,\dotsc,N$}
Initialize $\bm{A}^{(n)}\in\mathbb{R}^{I_n \times R}$, $n=1,\dotsc,N$, using random numbers\;
\Repeat{some termination criterion is reached}{
	(\textit{Supposing this is the $k$th iteration})\\
	\For{$n=1$ \KwTo $N$}{
		Make mode-$n$ unfolding of $\bm{\mathscr{X}}$ as $\bm{X}_{(n)}$\;
		Compute MTTKRP $\bm{X}_{(n)} \bm{B}_k^{(n)}$ and ${\big( \bm{B}_k^{(n)} \big)}^T \bm{B}_k^{(n)}$ based on \eqref{Eq:BtB}\;
		Update factor $\bm{A}_k^{(n)}$ using optimization method\;
	}
	Compute $\mathscr{O}_{\text{NCP}k}$ using \eqref{Eq:BasicNCPObjFast}, in which $n=N$\;
	Compute $\text{RelErr}_k$ using \eqref{Eq:RelErr2}\;
	Let the objective function of sparse NCP $\mathscr{O}_k=\mathscr{O}_{\text{NCP}k}$\;
	\For{$n=1$ \KwTo $N$}{
		$\displaystyle \mathscr{O}_k=\mathscr{O}_k + \frac{\alpha_n}{2}\norm{\bm{A}_k^{(n)}}_F^2 + \beta_n\sum_{r=1}^R\norm{\bm{a}_{rk}^{(n)}}_1$\;
	}
	\If{$\lvert \mathscr{O}_{k-1}-\mathscr{O}_{k} \rvert < \epsilon$}{
		Terminate iterating.
	}
	or\\
	\If{$\lvert \text{RelErr}_{k-1}-\text{RelErr}_k \rvert < \epsilon$}{
		Terminate iterating.
	}	
}
\end{algorithm}

For ANLS method, the \textbf{step 13} in \textbf{Algorithm \ref{Alg:ObjRelErr}} should be changed into
\begin{equation*}
\displaystyle \mathscr{O}_k=\mathscr{O}_k + \frac{\alpha_n}{2}\norm{\bm{A}_k^{(n)}}_F^2 + \frac{\beta_n}{2}\sum_{r=i}^{I_n}\norm{{\left[\bm{A}_k^{(n)} \right]}_{(i,:)} }_1^2
\end{equation*}
according to \eqref{Eq:ANLS1}.

\section{Experiments and Results}
We carried out the experiments on synthetic tensor data and real tensor data, both of which contain third-order and fourth-order cases. We compared the the abilities of sparse NCP methods to impose sparsity implemented by MU, ALS, HALS, APG, ANLS-AS and ANLS-BPP.

In all sparse NCP experiments, the factor matrices were initialized using nonnegative normally distributed random numbers by MATLAB function \texttt{max(0,randn($I_n,R$))}. We keep the Frobenius norm regularization in the sparse NCP model to make a fair comparison for all methods, especially the comparison of the objective function, although it is not necessary for some optimization methods, such as ANLS-AS, APG, HALS and MU. We set the Frobenius norm parameters by $\alpha_1=\alpha_2=, \dotsc, =\alpha_N=1e-6$. The $l_1$-norm regularization parameters of $\beta_n, n=1,\dotsc,N,$ in sparse NCP are the key elements to impose sparsity, which are the most important testing parameters in the experiments. In order to make it convenient to select and test the parameters, we also kept $\beta_n, n=1,\dotsc,N,$ the same in all modes of the tensor. After selecting the $\beta_n$, we calculated and evaluated the sparsity level \cite{Wang2018} of the factor matrices by
\begin{equation}
\text{Sparsity}_{\bm{A}^{(n)}}=\frac{\#\left\lbrace \bm{A}^{(n)}_{i,r}<T_s \right\rbrace}{I_n \times R},
\label{Eq:Sparsity}
\end{equation}
where $\#\left\lbrace \cdot \right\rbrace$ denotes the number of elements that are smaller than the threshold $T_s$ in factor matrix $\bm{A}^{(n)}$. In the experiments, we selected $T_s=1e-3$.

All the experiments were conducted on computer with Intel Core i5-4590 3.30GHz CPU, 8GB memory, 64-bit Windows 10 and MATLAB R2016b. The fundamental tensor computation was based on Tensor Toolbox 2.6 \cite{Bader2015,Bader2008,Bader2006}.

\subsection{Third-Order Synthetic Data}
\label{Section:Syn3rd}
\begin{figure}[!t]
\centering
\includegraphics[width=3.45in]{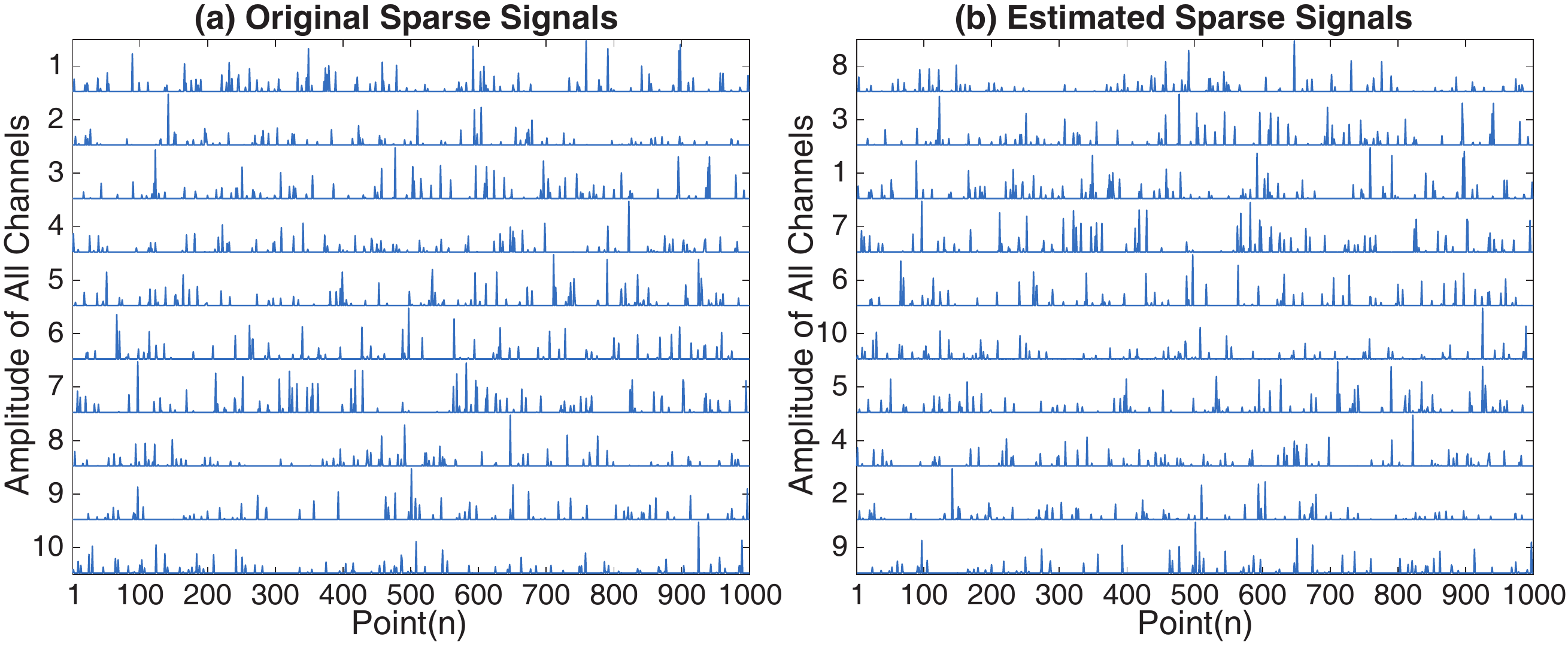}
\caption{Sparse and nonnegative signals used in synthetic tensor. (a) shows the original ten channels of signals. (b) shows the estimated ten channels of signals from third-order synthetic tensor by sparse NCP based on ANLS-BPP method, in which $\beta_n=0.5$.}
\label{Fig:Synthetic1SparseSignals}
\end{figure}

In the first experiment, we constructed a synthetic third-order tensor by 10 channels of simulated sparse and nonnegative signals\footnote{The sparse signals come from the file of \texttt{VSparse\_rand\_10.mat} included in NMFLAB, which can be downloaded from \href{http://www.bsp.brain.riken.jp/ICALAB/nmflab.html}{http://www.bsp.brain.riken.jp/ICALAB/nmflab.html}}, as shown in Fig. \ref{Fig:Synthetic1SparseSignals}(a). There are 1000 points in each channel, so the signal matrix is $\bm{S}^{(1)}=[\bm{s}_1,\dotsc,\bm{s}_{10}]\in\mathbb{R}^{1000 \times 10}$. Two uniformly distributed random matrices $\bm{A}^{(2)}, \bm{A}^{(3)} \in \mathbb{R}^{100 \times 10}$ were employed as mixing matrices, which were generated by \texttt{rand} function in MATLAB. Afterwards, we synthesized the third-order tensor by $\bm{\mathscr{X}}_{\text{Syn-3rd}}=\llbracket \bm{S}^{(1)},\bm{A}^{(2)},\bm{A}^{(3)} \rrbracket \in\mathbb{R}^{1000 \times 100 \times 100}$. Next, nonnegative Gaussian noise was added to the tensor with SNR of 40dB, which was generated by MATLAB code \texttt{max(0,randn(size($\bm{\mathscr{X}}$)))}.

\begin{table*}[!t]
\centering
\caption{Comparison of Sparse NCPs on Synthetic Tensor Data}
\label{Table:Synthetic3rdAnd4th}
\begin{IEEEeqnarraybox}[\IEEEeqnarraystrutmode\IEEEeqnarraystrutsizeadd{2pt}{0pt}]{x/x/x/x/x/x/x/x/v/x/x/x/x/x/x}
\IEEEeqnarraydblrulerowcut\\
&\IEEEeqnarraymulticol{7}{t}{Third-order Synthetic Tensor}&&\IEEEeqnarraymulticol{6}{t}{Fourth-order Synthetic Tensor}\\
\hfill\raisebox{-3pt}[0pt][0pt]{Algorithm}\hfill &\hfill\raisebox{-3pt}[0pt][0pt]{$\beta_n$}\hfill &\IEEEeqnarraymulticol{13}{h}{}\IEEEeqnarraystrutsize{0pt}{0pt}\\
&& \hfill Obj\hfill & \hfill RelErr\hfill & \hfill Time\hfill & \hfill Iter\hfill & \hfill Sparsity\hfill & \hfill Comp\hfill && \hfill Obj\hfill & \hfill RelErr\hfill & \hfill Time\hfill & \hfill Iter\hfill & \hfill Sparsity\hfill & \hfill Comp\hfill \IEEEeqnarraystrutsizeadd{0pt}{0pt}\\
\IEEEeqnarraydblrulerowcut\\
& 0 & 9.7243e+01 & 0.0082 & 8.6 & 168.1 & 0.6087 & 19.87 && 1.5094e+02 & 0.0082 & 87.8 & 43.7 & 0.6835 & 19.20\\
& 0.1 & 7.3466e+02 & 0.0083 & 14.1 & 357.3 & \textbf{0.9096} & \textbf{10.00} && 9.5641e+02 & 0.0084 & 92.0 & 45.6 & \textbf{0.8506} & \textbf{10.07}\\
& 0.5 & 2.1466e+03 & 0.0084 & 67.1 & 1713.7 & 0.7222 & \textbf{10.00} && 3.2866e+03 & 0.0085 & 412.2 & 204.8 & \textbf{0.8700} & \textbf{10.00}\\
\raisebox{1pt}[0pt][0pt]{ANLS} &\IEEEeqnarraymulticol{14}{t}{}\IEEEeqnarraystrutsize{0pt}{0pt}\\
\raisebox{-7pt}[0pt][0pt]{AS} &\IEEEeqnarraymulticol{14}{t}{}\IEEEeqnarraystrutsize{0pt}{0pt}\\
& 1 & 4.1733e+03 & 0.0084 & 50.0 & 1274.1 & 0.7198 & \textbf{10.00} && 4.3322e+03 & 0.0085 & 1043.8 & 520.2 & \textbf{0.8879} & \textbf{10.00}\\
& 2 & 8.2378e+03 & 0.0087 & 29.8 & 751.3 & 0.7211 & \textbf{10.00} && 6.4106e+03 & 0.0086 & 1283.3 & 641.8 & \textbf{0.8896} & \textbf{10.00}\\
& 3 & 1.2298e+04 & 0.0092 & 21.7 & 546.0 & 0.7235 & \textbf{10.00} && 8.3929e+03 & 0.0087 & 1547.8 & 768.0 & \textbf{0.8673} & \textbf{10.00}\\
\hline
& 0 & 9.7240e+01 & 0.0082 & 8.2 & 165.6 & 0.6173 & 19.90 && 1.5095e+02 & 0.0082 & 99.5 & 49.8 & 0.6848 & 19.23\\
& 0.1 & 7.3447e+02 & 0.0083 & 16.4 & 390.0 & \textbf{0.9045} & \textbf{10.00} && 9.1615e+02 & 0.0085 & 82.3 & 41.0 & \textbf{0.8658} & \textbf{10.23}\\
& 0.5 & 2.1475e+03 & 0.0084 & 70.8 & 1735.1 & 0.7228 & \textbf{10.00} && 3.2693e+03 & 0.0085 & 362.1 & 181.0 & \textbf{0.8679} & \textbf{10.00}\\
\raisebox{1pt}[0pt][0pt]{ANLS} &\IEEEeqnarraymulticol{7}{t}{}\IEEEeqnarraystrutsize{0pt}{0pt}\\
\raisebox{-7pt}[0pt][0pt]{BPP} &\IEEEeqnarraymulticol{7}{t}{}\IEEEeqnarraystrutsize{0pt}{0pt}\\
& 1 & 4.1734e+03 & 0.0084 & 52.0 & 1269.6 & 0.7195 & \textbf{10.00} && 4.4226e+03 & 0.0085 & 1116.8 & 559.7 & \textbf{0.8716} & \textbf{10.00}\\
& 2 & 8.2374e+03 & 0.0087 & 31.5 & 762.4 & 0.7213 & \textbf{10.00} && 6.5015e+03 & 0.0086 & 1192.1 & 596.8 & \textbf{0.8886} & \textbf{10.00}\\
& 3 & 1.2298e+04 & 0.0092 & 22.4 & 540.3 & 0.7237 & \textbf{10.00} && 8.3143e+03 & 0.0087 & 1594.6 & 795.6 & \textbf{0.8704} & \textbf{10.00}\\
\hline
& 0 & 9.7435e+01 & 0.0082 & 37.5 & 992.0 & 0.4195 & 20.00 && 1.5378e+02 & 0.0083 & 1292.9 & 647.1 & 0.1841 & 20.00\\
& 0.1 & 5.0500e+02 & 0.0083 & 12.0 & 321.6 & 0.5187 & 19.83 && 6.9398e+02 & 0.0085 & 448.6 & 224.8 & 0.5522 & 15.50\\
& 0.5 & 1.7657e+03 & 0.0084 & 12.5 & 334.3 & 0.6363 & 17.83 && 2.4718e+03 & 0.0085 & 421.3 & 211.1 & 0.6549 & 12.70\\
\raisebox{-3pt}[0pt][0pt]{APG} &\IEEEeqnarraymulticol{7}{t}{}\IEEEeqnarraystrutsize{0pt}{0pt}\\
& 1 & 2.9607e+03 & 0.0084 & 25.5 & 667.3 & 0.7365 & 14.97 && 4.4068e+03 & 0.0086 & 500.7 & 249.7 & 0.7092 & 12.13\\
& 2 & 4.2958e+03 & 0.0084 & 68.2 & 1824.5 & \textbf{0.8980} & \textbf{10.67} && 6.4829e+03 & 0.0087 & 969.8 & 486.4 & \textbf{0.7863} & \textbf{10.97}\\
& 3 & 6.2480e+03 & 0.0085 & 60.9 & 1632.1 & \textbf{0.9086} & \textbf{10.73} && 8.1431e+03 & 0.0088 & 1106.6 & 557.3 & \textbf{0.8270} & \textbf{10.67}\\
\hline
& 0 & 9.8720e+01 & 0.0083 & 180 & 4976.2 & 0.7270 & 20.00 && 1.6924e+02 & 0.0087 & 1800 & 902.2 & 0.7072 & 20.00\\
& 0.1 & 3.5118e+02 & 0.0084 & 158.4 & 4364.9 & 0.8435 & 17.07 && 4.6624e+02 & 0.0087 & 1800 & 898.4 & 0.7709 & 19.77\\
& 0.5 & 1.1763e+03 & 0.0084 & 134.5 & 3704.6 & 0.8726 & 13.00 && 1.4612e+03 & 0.0087 & 1800 & 897.4 & 0.8410 & 16.30\\
\raisebox{-3pt}[0pt][0pt]{MU} &\IEEEeqnarraymulticol{7}{t}{}\IEEEeqnarraystrutsize{0pt}{0pt}\\
& 1 & 2.1223e+03 & 0.0084 & 131.7 & 3625.6 & 0.8728 & 11.33 && 2.5216e+03 & 0.0087 & 1800 & 904.3 & 0.8722 & 14.57\\
& 2 & 3.9818e+03 & 0.0084 & 134.7 & 3720.2 & 0.8720 & \textbf{10.50} && 4.2640e+03 & 0.0087 & 1800 & 897.8 & 0.8973 & 12.80\\
& 3 & 5.9273e+03 & 0.0085 & 123.8 & 3416.9 & 0.8640 & \textbf{10.43} && 5.9319e+03 & 0.0088 & 1800 & 898.6 & 0.9022 & 12.20\\
\hline
& 0 & 9.8608e+01 & 0.0082 & 25.7 & 670.6 & 0.8453 & 20.00 && 1.5537e+02 & 0.0083 & 491.6 & 244.6 & 0.8636 & 20.00\\
& 0.1 & 4.1446e+03 & 0.0099 & 10.6 & 274.7 & 0.7929 & 20.00 && 1.3866e+03 & 0.0084 & 442.5 & 220.1 & 0.8331 & 20.00\\
& 0.5 & 1.9432e+04 & 0.0222 & 17.7 & 460.1 & 0.6651 & 20.00 && 6.2725e+03 & 0.0089 & 678.6 & 339.8 & 0.7907 & 19.53\\
\raisebox{-3pt}[0pt][0pt]{HALS} &\IEEEeqnarraymulticol{7}{t}{}\IEEEeqnarraystrutsize{0pt}{0pt}\\
& 1 & 3.7638e+04 & 0.0364 & 16.0 & 413.3 & 0.6034 & 20.00 && 1.2357e+04 & 0.0102 & 367.1 & 183.4 & 0.7512 & 19.57\\
& 2 & 7.2050e+04 & 0.0603 & 13.0 & 338.3 & 0.5462 & 19.93 && 2.4423e+04 & 0.0141 & 223.1 & 111.3 & 0.7180 & 19.33\\
& 3 & 1.0444e+05 & 0.0812 & 14.0 & 362.8 & 0.5130 & 19.97 && 3.6296e+04 & 0.0184 & 227.8 & 113.0 & 0.7168 & 19.10\\
\hline
& 0 & 1.0115e+02 & 0.0083 & 9.2 & 234.5 & 0.6350 & 16.73 && 1.6167e+02 & 0.0083 & 152.7 & 76.1 & 0.6418 & 17.00\\
& 0.1 & 5.5442e+02 & 0.0085 & 14.7 & 405.0 & \textbf{0.9278} & \textbf{10.20} && 5.6488e+03 & 0.0163 & 657.5 & 327.4 & \textbf{0.7884} & \textbf{10.30}\\
& 0.5 & 1.5598e+03 & 0.0083 & 44.9 & 1228.3 & \textbf{0.9157} & \textbf{10.03} && 1.0572e+04 & 0.0164 & 926.7 & 467.1 & \textbf{0.8084} & \textbf{10.10}\\
\raisebox{-3pt}[0pt][0pt]{ALS} &\IEEEeqnarraymulticol{7}{t}{}\IEEEeqnarraystrutsize{0pt}{0pt}\\
& 1 & 2.2705e+03 & 0.0084 & 88.9 & 2416.1 & \textbf{0.8938} & \textbf{10.00} && 1.1493e+04 & 0.0171 & 1244.0 & 621.5 & \textbf{0.8050} & \textbf{9.93}\\
& 2 & 4.5176e+03 & 0.0085 & 90.0 & 2463.6 & \textbf{0.8896} & \textbf{10.00} && 1.8490e+04 & 0.0241 & 1160.1 & 579.9 & \textbf{0.8090} & \textbf{9.90}\\
& 3 & 6.6982e+03 & 0.0085 & 79.4 & 2177.8 & \textbf{0.8810} & \textbf{10.00} && 2.2452e+04 & 0.0256 & 1187.6 & 593.5 & \textbf{0.8472} & \textbf{9.87}\\
\IEEEeqnarraydblrulerowcut\\
\IEEEeqnarraymulticol{15}{s}{\textbf{Note:} For the third-order synthetic tensor, the threshold of stopping condition is 1e-8 based on relative error change, and the}\\
\IEEEeqnarraymulticol{15}{s}{maximum running time is 180s. For the fourth-order synthetic tensor, the threshold is 1e-7, and the maximum running time}\\
\IEEEeqnarraymulticol{15}{s}{is 1800s. The values are the average after 30 times of running.}
\end{IEEEeqnarraybox}
\end{table*}

\begin{figure*}[!t]
\centering
\includegraphics[width=0.8\linewidth]{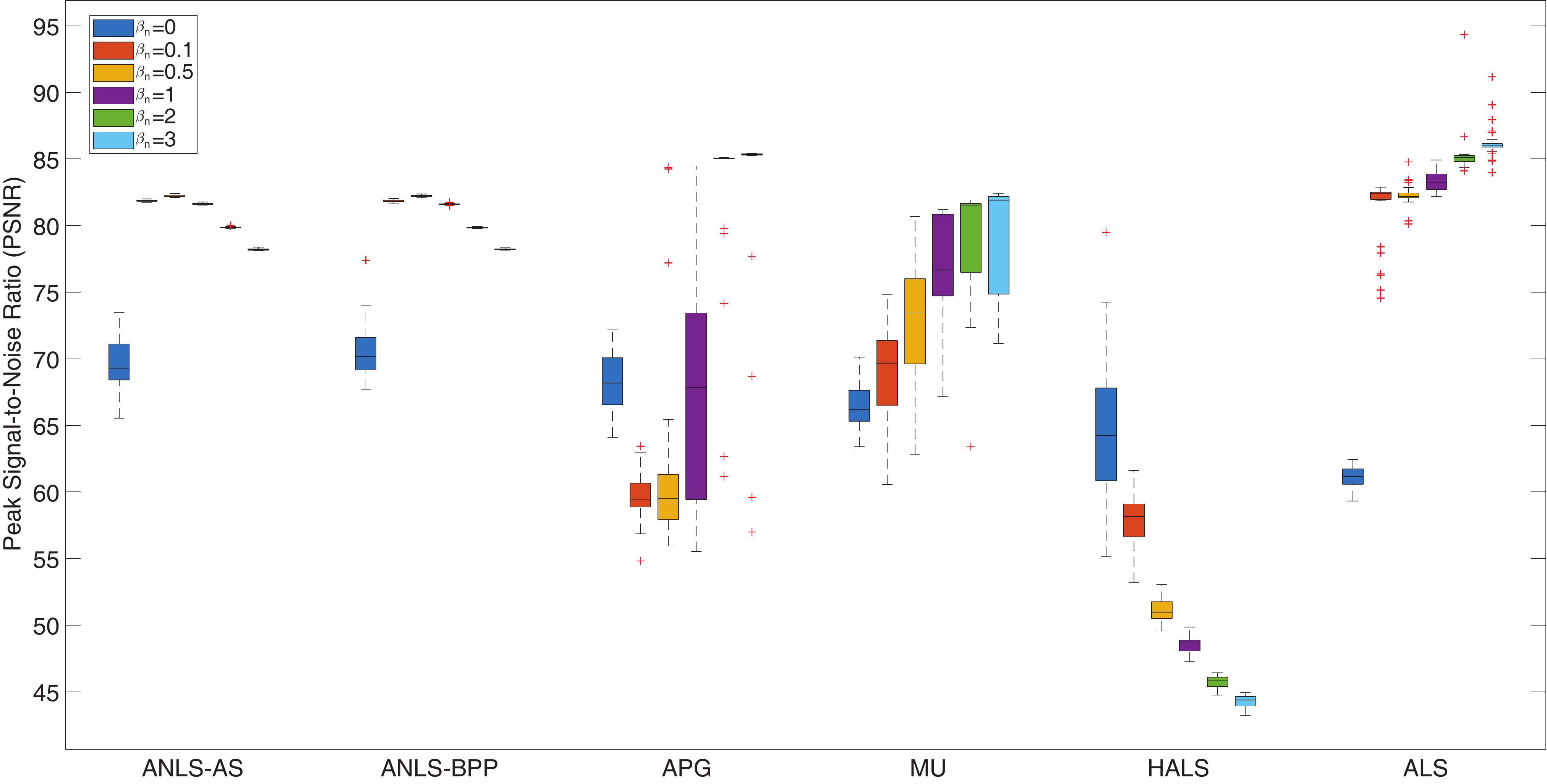}
\caption{Boxplot of Peak Signal-to-Noise Ratio (PSNR) for sparse NCP methods on third-order synthetic tensor. The PSNR is a measure of accuracy of the estimated sparse signals compared with the true signals.}
\label{Fig:Synthetic3rdBoxplot}
\end{figure*}

\begin{figure*}[!t]
\centering
\includegraphics[width=\linewidth]{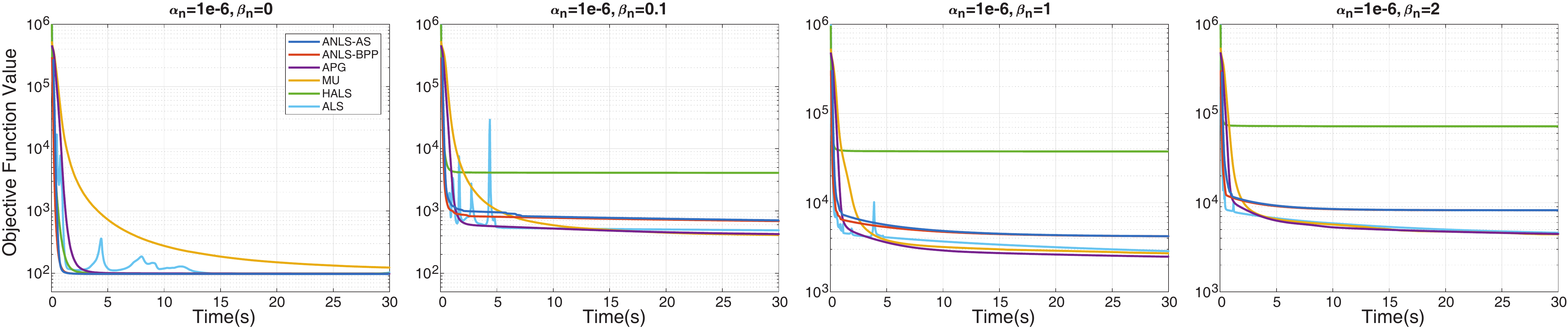}
\caption{The Objective Function Value Curves of Sparse NCPs on Third-order Synthetic Tensor With Fixed Time Limit of 30s.}
\label{Fig:Synthetic3rdObj}
\end{figure*}

For all sparse NCP methods, we used the stop condition of relative error change in \eqref{Eq:StopRelErr}, in which the threshold is $\epsilon=1e-8$. The maximum running time was set by $180$ seconds. We selected a larger value of 20 as the number of components for tensor decomposition\footnote{Since 10 channels of signals are mixed in the tensor, naturally, 10 should be selected as the optimal component number. The number of components might also be estimated by some classical methods, such as DIFFIT \cite{Timmerman2000}. However, we selected 20 according to the purpose of the experiment.}. The reason is that we intend to recover the 10 channels of true signal just by imposing sparse regularization during decomposition, even though we don't know the exact optimal number of components. We selected values of $\beta_n=0,0.1,0.5,1,2,3$ for all the optimization methods to evaluate their abilities to impose sparsity. The selection of sparse regularization parameters depends on the tensor data. After tensor decomposition, the values of objective function value, relative error, running time, iteration number, the sparsity level and nonzero component number of the estimated signal factor matrix were recorded as the performance evaluation criteria. For all optimization methods with each $\beta_n$, the sparse NCP was run 30 times, and the average values of all criteria were computed. The results are shown in \textbf{Table \ref{Table:Synthetic3rdAnd4th}}.

From \textbf{Table \ref{Table:Synthetic3rdAnd4th}}, it can be found that all ANLS-AS, ANLS-BPP, APG, MU, ALS methods can impose sparsity with proper sparse regularization parameter $\beta_n$. With certain sparse regularization, 10 nonzero components are retained in the mode-1 factor matrix, which represent the 10 channels of sparse signals extracted from the mixed tensor. ANLS-AS, ANLS-BPP, APG, and ALS exhibit comparatively low relative errors and less running time, whereas MU converges very slowly. On the other hand, HALS fails to impose sparsity even with a larger $\beta_n$ and can hardly reduce the number of nonzero components. HALS also has very large objective function value and high relative error with larger $\beta_n$.

It can be inferred from \textbf{Table \ref{Table:Synthetic3rdAnd4th}} that, after properly tuning the sparse regularization parameter $\beta_n$, weak components will be removed (set to 0), weak elements in strong components will be prohibited, and the true 10 channels of sparse signals will be recovered. One of the recovered sparse signal matrix by ANLS-BPP is shown in Fig. \ref{Fig:Synthetic1SparseSignals}(b). Afterwards, the accurary of the recovered signals should be evaluated. Let $\bm{T}^{(1)}=[\bm{t}_1,\dotsc,\bm{t}_{\tilde{R}}]\in\mathbb{R}^{L \times \tilde{R}}$ represent the estimated matrix of sparse signals, in which $\tilde{R}$ is the number of nonzero components and $L$ is the length of component ($L=1000$ in this experiment). We evaluate the accuracy of the estimated matrix $\bm{T}^{(1)}$ compared with original sparse signals $\bm{S}^{(1)}$ by Peak Signal-to-Noise-Ratio (PSNR, see Chapter 3 in \cite{Cichocki2009})
\begin{equation}
\text{PSNR}=\frac{1}{\tilde{R}}\sum_{r=1}^{\tilde{R}}10\text{log}_{10} \left( \frac{L}{\norm{\hat{\bm{t}}_r-\hat{\bm{s}}_c}_2^2}  \right),
\end{equation}
where $\hat{\bm{t}}_r$ is the $r$th normalized estimated sparse signal, and $\hat{\bm{s}}_c$ is the normalized reference sparse signal. $\hat{\bm{s}}_c$ comes from $\bm{S}^{(1)}$, which has the highest correlation coefficient with $\hat{\bm{t}}_r$. For all the optimization methods with each $\beta_n$, all the PSNR values were recorded after 30 times of running of sparse NCP. Subsequently, box plot of PSNR for MU, ALS, HALS, APG, ANLS-AS, ANLS-BPP was drawn in \textbf{Fig. \ref{Fig:Synthetic3rdBoxplot}}.

In \textbf{Fig. \ref{Fig:Synthetic3rdBoxplot}}, it is clear that APG and ALS methods have higher PSNR at larger sparse regularization parameters of $\beta_n=2,3$, with which they recover the 10 channels of sparse signals more precisely. For ANLS-AS and ANLS-BPP, their PSNR values decrease when $\beta_n>0.5$. The reason might be that ANLS framework uses the squared $l_1$-norm as the sparse regularization item, which is more sensitive to the sparse regularization parameter $\beta_n$ (smaller $\beta_n$ will cause higher sparsity, and larger $\beta_n$ might spoil the decomposition). \textbf{Fig. \ref{Fig:Synthetic3rdBoxplot}} also shows that HALS performs poorly with sparse sparse regularization.

We also recorded the objective function values of all sparse NCP methods within the first 30 seconds. The results of these methods with $\beta_n=0,0.1,1,2$ are shown in \textbf{Fig. \ref{Fig:Synthetic3rdObj}}. In the results, we find that MU converges very slowly in all cases, but it can minimize the objective function to a low level gradually. ALS method is not stable, which sometimes can't ensure that the objective function decreases. However, the stability of ALS will be improved with higher sparse regularization parameters. HALS runs much fast, but its objective function becomes very large when sparse regularization is added. APG shows excellent performances in both running speed and the ability to minimize the objective function in the third-order tensor case. ANLS-AS and ANLS-BPP can minimize the objective function very fast at the beginning, but their objective function values are a bit higher when sparse regularization is added. This is due to the squared $l_1$-norm item in ANLS framework.

\begin{figure*}[!t]
\centering
\includegraphics[width=0.8\linewidth]{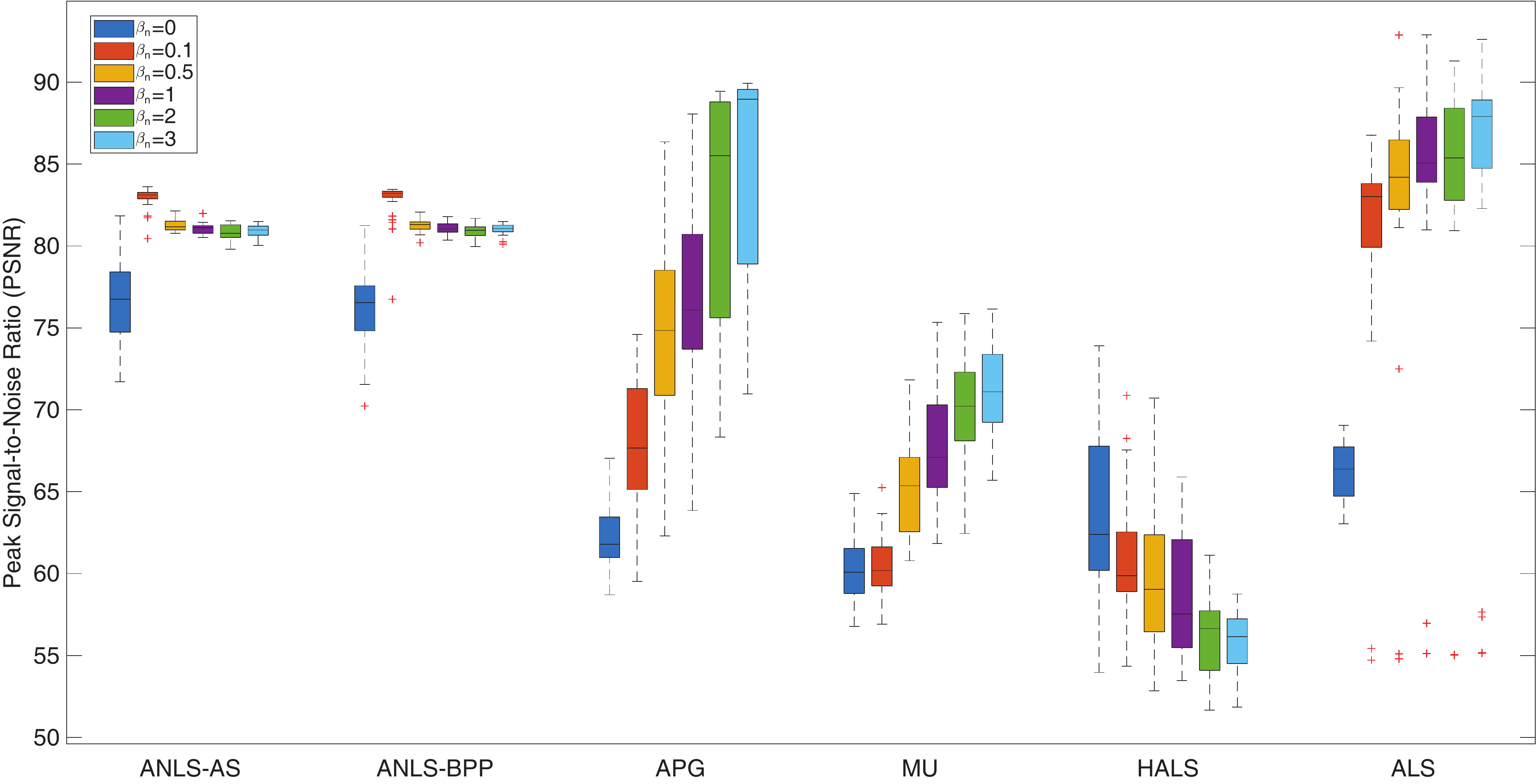}
\caption{Boxplot of Peak Signal-to-Noise Ratio (PSNR) for sparse NCP methods on Fourth-order synthetic tensor. The PSNR is a measure of accuracy of the estimated sparse signals compared with the true signals.}
\label{Fig:Synthetic4thBoxplot}
\end{figure*}

\begin{figure*}[!t]
\centering
\includegraphics[width=\linewidth]{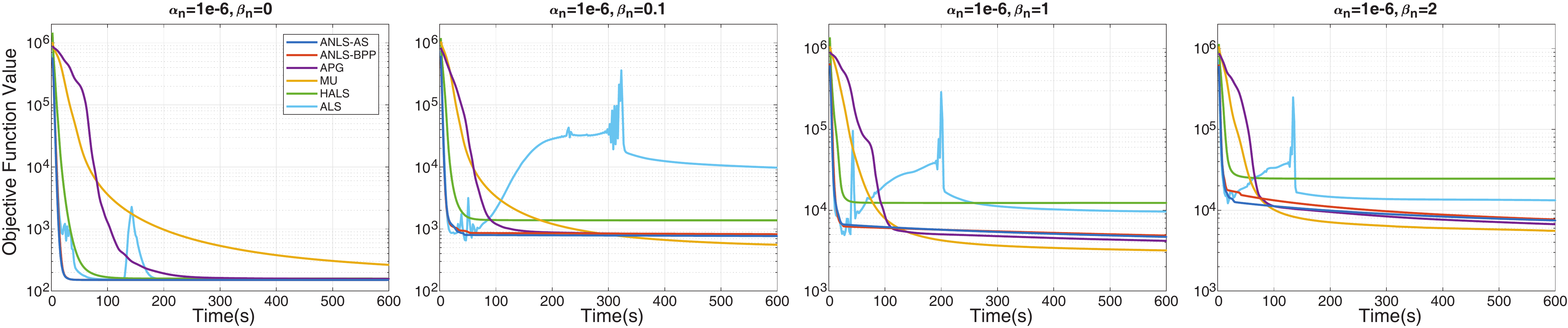}
\caption{The Objective Function Value Curves of Sparse NCPs on Fourth-order Synthetic Tensor With Fixed Time Limit of 600s.}
\label{Fig:Synthetic4thObj}
\end{figure*}

\subsection{Fourth-Order Synthetic Data}
\label{Section:Syn4th}
In the second experiment, we synthesized a fourth-order tensor by the same 10 channels of simulated sparse and nonnegative signals as that in last section. The tensor was synthesized by $\bm{\mathscr{X}}_{\text{Syn-4th}}=\llbracket \bm{S}^{(1)},\bm{A}^{(2)},\bm{A}^{(3)},\bm{A}^{(4)} \rrbracket \in\mathbb{R}^{1000 \times 100 \times 100 \times 5}$, in which $\bm{S}^{(1)}$ is the simulated sparse signals, and $\bm{A}^{(2)},\bm{A}^{(3)} \in \mathbb{R}^{100 \times 10},\bm{A}^{(4)} \in \mathbb{R}^{5 \times 10}$ are random matrices in uniform distribution. Nonnegative Gaussian noise was added to this fourth-order tensor with SNR of 40dB.

Processing higher-order (order $\geqslant 4$) is very challenging due to the huge amount of data and the high complexity of optimization method. In this fourth-order tensor experiment, we selected $\epsilon=1e-7$ as the threshold of stopping condition based on the relative error change in \eqref{Eq:StopRelErr}, and set 1800 seconds as the maximum running time. Other parameter settings and experimental methods are the same as those in subsection \ref{Section:Syn3rd}. The averages of objective function, relative error, running time, iteration number, the sparsity level and nonzero component number of the estimated signal factor matrix after 30 times of running are recorded in \textbf{Table \ref{Table:Synthetic3rdAnd4th}}. The box plot of PSNR for all methods after 30 times of run is shown in \textbf{Fig. \ref{Fig:Synthetic4thBoxplot}}. The results of the fourth-order synthetic tensor decomposition in \textbf{Table \ref{Table:Synthetic3rdAnd4th}} and \textbf{Fig. \ref{Fig:Synthetic4thBoxplot}} have proven once more the effectiveness of MU, ALS, APG, ANLS-AS, ANLS-BPP to impose sparsity and estimate the true sparse signals.

We recorded the objective function values of all sparse NCP methods for the fourth-order tensor within the first 600 seconds. The results of these methods with $\beta_n=0,0.1,1,2$ are shown in \textbf{Fig. \ref{Fig:Synthetic4thObj}}. It is clear to see that ALS method is still not stable, which can't guarantee the objective function to decrease. MU method still shows slow convergence. Surprisingly, the APG method that presents excellent performance for third-order tensor runs very slowly for the fourth-order tensor, which performs even more poorly than MU at the first tens of seconds. HALS has very large objective function value when sparse regularization is imposed. However, both ANLS-AS and ANLS-BPP still converge very fast for the fourth-order sparse NCP. When sparse regularization is imposed, ANLS-AS and ANLS-BPP have slightly higher objective function values due to the squared $l_1$-norm item in ANLS framework.

\begin{table*}[!t]
\centering
\caption{Comparison of Sparse NCPs on ERP Tensor Data}
\label{Table:ERP3rdAnd4th}
\begin{IEEEeqnarraybox}[\IEEEeqnarraystrutmode\IEEEeqnarraystrutsizeadd{2pt}{0pt}]{x/x/x/x/x/x/x/x/v/x/x/x/x/x/x}
\IEEEeqnarraydblrulerowcut\\
&\IEEEeqnarraymulticol{7}{t}{Third-order ERP Tensor}&&\IEEEeqnarraymulticol{6}{t}{Fourth-order ERP Tensor}\\
\hfill\raisebox{-3pt}[0pt][0pt]{Algorithm}\hfill &\hfill\raisebox{-3pt}[0pt][0pt]{$\beta_n$}\hfill &\IEEEeqnarraymulticol{13}{h}{}\IEEEeqnarraystrutsize{0pt}{0pt}\\
&& \hfill Obj\hfill & \hfill RelErr\hfill & \hfill Time\hfill & \hfill Iter\hfill & \hfill Sparsity\hfill & \hfill Comp\hfill && \hfill Obj\hfill & \hfill RelErr\hfill & \hfill Time\hfill & \hfill Iter\hfill & \hfill Sparsity\hfill & \hfill Comp\hfill \IEEEeqnarraystrutsizeadd{0pt}{0pt}\\
\IEEEeqnarraydblrulerowcut\\
& 0 & 3.2211e+05 & 0.0916 & 119.5 & 1410.9 & \textbf{0.2594} & 38.87 && 5.8001e+05 & 0.1229 & 18.6 & 608.5 & \textbf{0.3338} & 34.17\\
& 10 & 1.1203e+06 & 0.1439 & 44.3 & 687.9 & \textbf{0.5645} & 20.80 && 9.9992e+05 & 0.1360 & 20.2 & 686.8 & \textbf{0.4283} & 29.60\\
& 50 & 2.2425e+06 & 0.1989 & 33.7 & 549.3 & \textbf{0.7416} & 11.53 && 1.8522e+06 & 0.1655 & 16.1 & 556.7 & \textbf{0.5640} & 21.77\\
\raisebox{1pt}[0pt][0pt]{ANLS} &\IEEEeqnarraymulticol{14}{t}{}\IEEEeqnarraystrutsize{0pt}{0pt}\\
\raisebox{-7pt}[0pt][0pt]{AS} &\IEEEeqnarraymulticol{14}{t}{}\IEEEeqnarraystrutsize{0pt}{0pt}\\
& 100 & 3.0417e+06 & 0.2376 & 25.3 & 413.7 & \textbf{0.8200} & 7.83 && 2.4220e+06 & 0.1891 & 13.5 & 491.2 & \textbf{0.6521} & 16.43\\
& 200 & 3.8588e+06 & 0.2731 & 24.0 & 397.0 & \textbf{0.8774} & 5.27 && 3.1446e+06 & 0.2185 & 12.1 & 456.1 & \textbf{0.7380} & 11.90\\
& 300 & 4.3772e+06 & 0.2963 & 17.7 & 290.3 & \textbf{0.9071} & 3.93 && 3.6604e+06 & 0.2346 & 11.9 & 438.5 & \textbf{0.7715} & 10.33\\
\hline
& 0 & 3.1067e+05 & 0.0900 & 119.7 & 1498.7 & \textbf{0.2509} & 39.70 && 5.6736e+05 & 0.1216 & 23.4 & 713.4 & \textbf{0.3247} & 34.90\\
& 10 & 1.1248e+06 & 0.1451 & 40.9 & 691.9 & \textbf{0.5740} & 20.37 && 9.9176e+05 & 0.1346 & 20.0 & 694.1 & \textbf{0.4230} & 30.00\\
& 50 & 2.2574e+06 & 0.2015 & 26.5 & 479.7 & \textbf{0.7480} & 11.17 && 1.8548e+06 & 0.1651 & 14.6 & 517.0 & \textbf{0.5660} & 21.77\\
\raisebox{1pt}[0pt][0pt]{ANLS} &\IEEEeqnarraymulticol{7}{t}{}\IEEEeqnarraystrutsize{0pt}{0pt}\\
\raisebox{-7pt}[0pt][0pt]{BPP} &\IEEEeqnarraymulticol{7}{t}{}\IEEEeqnarraystrutsize{0pt}{0pt}\\
& 100 & 3.0346e+06 & 0.2350 & 26.5 & 480.7 & \textbf{0.8146} & 8.10 && 2.4312e+06 & 0.1899 & 13.2 & 469.8 & \textbf{0.6560} & 16.37\\
& 200 & 3.8466e+06 & 0.2715 & 27.1 & 507.1 & \textbf{0.8743} & 5.47 && 3.1594e+06 & 0.2158 & 9.8 & 362.4 & \textbf{0.7294} & 12.23\\
& 300 & 4.3186e+06 & 0.2914 & 21.7 & 406.0 & \textbf{0.9020} & 4.20 && 3.6794e+06 & 0.2350 & 13.3 & 480.5 & \textbf{0.7690} & 10.43\\
\hline
& 0 & 3.0688e+05 & 0.0895 & 134.4 & 2540.1 & \textbf{0.2602} & 40.00 && 4.8108e+05 & 0.1120 & 99.1 & 2113.5 & \textbf{0.2468} & 40.00\\
& 10 & 3.8277e+05 & 0.0897 & 119.9 & 2247.9 & \textbf{0.2861} & 40.00 && 6.6882e+05 & 0.1263 & 71.5 & 1525.3 & \textbf{0.3509} & 33.03\\
& 50 & 6.6367e+05 & 0.0919 & 89.1 & 1678.9 & \textbf{0.3160} & 39.03 && 1.2169e+06 & 0.1651 & 36.2 & 775.0 & \textbf{0.5748} & 20.73\\
\raisebox{-3pt}[0pt][0pt]{APG} &\IEEEeqnarraymulticol{7}{t}{}\IEEEeqnarraystrutsize{0pt}{0pt}\\
& 100 & 1.0160e+06 & 0.1004 & 54.8 & 1038.4 & \textbf{0.3783} & 35.67 && 1.6803e+06 & 0.1936 & 30.1 & 655.7 & \textbf{0.6793} & 15.03\\
& 200 & 1.6074e+06 & 0.1226 & 43.2 & 824.2 & \textbf{0.5075} & 27.33 && 2.5051e+06 & 0.2367 & 22.3 & 486.2 & \textbf{0.7860} & 9.60\\
& 300 & 2.1091e+06 & 0.1447 & 37.0 & 708.7 & \textbf{0.6003} & 21.23 && 2.9254e+06 & 0.2555 & 20.2 & 437.8 & \textbf{0.8217} & 7.73\\
\hline
& 0 & 3.3865e+05 & 0.0940 & 240 & 5165.7 & 0.2466 & 40.00 && 5.0307e+05 & 0.1145 & 120 & 4490.9 & 0.2408 & 40.00\\
& 10 & 4.1809e+05 & 0.0950 & 240 & 4934.4 & 0.3004 & 40.00 && 5.4429e+05 & 0.1149 & 120 & 4478.4 & 0.2445 & 40.00\\
& 50 & 6.9031e+05 & 0.0951 & 240 & 5070.4 & 0.3079 & 39.80 && 6.7206e+05 & 0.1147 & 120 & 4470.2 & 0.2502 & 40.00\\
\raisebox{-3pt}[0pt][0pt]{MU} &\IEEEeqnarraymulticol{7}{t}{}\IEEEeqnarraystrutsize{0pt}{0pt}\\
& 100 & 1.0382e+06 & 0.0991 & 240 & 5035.2 & 0.3374 & 38.77 && 8.3825e+05 & 0.1148 & 120 & 4706.8 & 0.2501 & 39.97\\
& 200 & 1.6559e+06 & 0.1150 & 153.3 & 3090.9 & 0.4366 & 33.33 && 1.1797e+06 & 0.1170 & 119.2 & 4697.8 & 0.2637 & 39.53\\
& 300 & 2.1959e+06 & 0.1296 & 73.0 & 1588.1 & 0.5004 & 29.27 && 1.5014e+06 & 0.1207 & 111.3 & 4351.7 & 0.2949 & 38.07\\
\hline
& 0 & 3.0571e+05 & 0.0893 & 139.8 & 2864.5 & 0.3287 & 40.00 && 4.8047e+05 & 0.1119 & 48.5 & 1780.7 & 0.2458 & 40.00\\
& 10 & 1.1920e+06 & 0.1075 & 28.6 & 579.5 & 0.1869 & 40.00 && 1.4878e+06 & 0.1318 & 16.1 & 593.7 & 0.1784 & 40.00\\
& 50 & 3.7030e+06 & 0.1479 & 29.3 & 592.4 & 0.0683 & 40.00 && 4.1396e+06 & 0.1779 & 15.1 & 555.7 & 0.0855 & 40.00\\
\raisebox{-3pt}[0pt][0pt]{HALS} &\IEEEeqnarraymulticol{7}{t}{}\IEEEeqnarraystrutsize{0pt}{0pt}\\
& 100 & 6.3899e+06 & 0.1845 & 25.6 & 521.5 & 0.0409 & 40.00 && 6.8072e+06 & 0.2177 & 17.2 & 622.6 & 0.0377 & 40.00\\
& 200 & 1.1155e+07 & 0.2471 & 91.1 & 1859.7 & 0.1135 & 35.93 && 1.1392e+07 & 0.2763 & 33.6 & 1219.8 & 0.0602 & 38.17\\
& 300 & 1.5526e+07 & 0.3137 & 175.1 & 3510.1 & 0.2825 & 29.00 && 1.5458e+07 & 0.3332 & 10.1 & 371.8 & 0.1817 & 33.07\\
\hline
& 0 & 1.4336e+06 & 0.1822 & 237.6 & 4079.6 & 0.6615 & 20.97 && 1.6875e+06 & 0.1474 & 115.1 & 2266.5 & 0.5117 & 26.27\\
& 10 & 5.4467e+06 & 0.1706 & 228.2 & 4219.0 & 0.7123 & 18.63 && 7.1430e+06 & 0.1619 & 119.0 & 2465.7 & 0.5151 & 27.07\\
& 50 & 6.0430e+06 & 0.2058 & 140.6 & 2711.0 & 0.9208 & 5.53 && 7.3298e+06 & 0.1764 & 114.9 & 2507.3 & 0.5579 & 24.23\\
\raisebox{-3pt}[0pt][0pt]{ALS} &\IEEEeqnarraymulticol{7}{t}{}\IEEEeqnarraystrutsize{0pt}{0pt}\\
& 100 & 8.4312e+06 & 0.3773 & 21.7 & 459.6 & 0.9454 & 2.83 && 7.7009e+06 & 0.1738 & 116.5 & 2955.9 & 0.5582 & 24.13\\
& 200 & 2.3166e+07 & 0.7066 & 12.4 & 275.5 & 0.9817 & 0.87 && 7.0916e+06 & 0.1899 & 95.9 & 2605.2 & 0.6020 & 20.77\\
& 300 & 3.5148e+07 & 0.9392 & 2.0 & 44.0 & 0.9968 & 0.13 && 7.4508e+06 & 0.1955 & 83.3 & 2445.7 & 0.6307 & 18.83\\
\IEEEeqnarraydblrulerowcut\\
\IEEEeqnarraymulticol{15}{s}{\textbf{Note:} The threshold of stopping condition is 1e-8 based on relative error change. For the third-order tensor, the maximum running}\\
\IEEEeqnarraymulticol{15}{s}{time is 240s, and the sparsity and nonzero component number are computed based on the frequency-time factor matrix. For the}\\
\IEEEeqnarraymulticol{15}{s}{fourth-order tensor, the maximum running time is 120s, and the sparsity and nonzero component number are computed based on}\\
\IEEEeqnarraymulticol{15}{s}{the temporal factor matrix. The values in the table are the average after 30 times of running.}\\
\IEEEeqnarraymulticol{15}{s}{ }\\
\end{IEEEeqnarraybox}
\end{table*}

\subsection{Third-Order ERP Data}
\label{Section:ERP3rd}
In the third experiment, we utilized an open preprocessed ERP tensor\footnote{Data website: \href{http://www.escience.cn/people/cong/AdvancedSP\_ERP.html}{http://www.escience.cn/people/cong/AdvancedSP\_ERP.html}}. Original data are organized in fourth-order form, whose size is channel \texttimes~frequency \texttimes~time \texttimes~subject-group = 9 \texttimes~71 \texttimes~60 \texttimes~42. The 9 channel points denote the 9 electrodes on the scalp, the 71 frequency points show the spectrum within 1-15Hz, the 60 time points illustrate the temporal energy between 0-300ms, and the 42 subject-group points include 21 subjects with reading disability (RD) and 21 subjects with attention deficit (AD) \cite{Cong2012}. In this experiment, we merged the modes of frequency and time, by which the original tensor was reshaped into a third-order tensor with size channel \texttimes~frequency-time \texttimes~subject-group = 9 \texttimes~4260 \texttimes~ 42.

For this third-order tensor, the number of components is set by 40 according to previous study \cite{Cong2012}. For all methods, the threshold of stopping condition was set by 1e-8 based on the relative error change, and the maximum running time was 240s. The values of $\beta_n=0,10,50,100,200,300$ were tested for all methods. We recorded the objective function, relative error, running time, iteration number, the sparsity level and nonzero component number of the frequency-time factor matrix for all methods. The average values after 30 times of run are recorded in \textbf{Table \ref{Table:ERP3rdAnd4th}}.

It can be found from \textbf{Table \ref{Table:ERP3rdAnd4th}} that ANLS-AS, ANLS-BPP and APG are very effective to impose sparsity on the factor matrix and reduce components number by adjusting $\beta_n$, which costs comparatively less time. MU is not very effective to impose sparsity, which often reaches the limit of running time with slow convergence. Both HALS and ALS fail to impose proper sparsity by adjusting $\beta_n$, which also show high objective function values and relative errors with large $\beta_n$.

\begin{figure*}[!t]
\centering
\includegraphics[width=\linewidth]{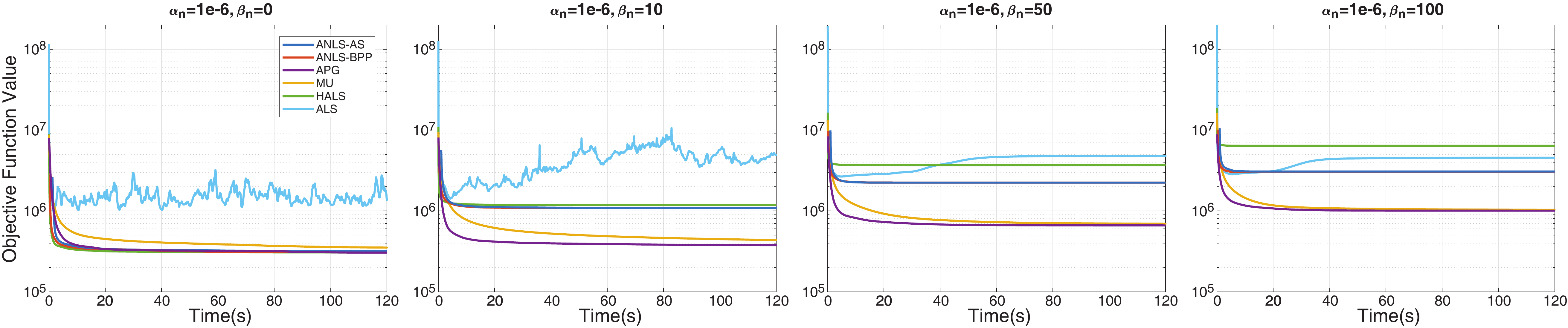}
\caption{The Objective Function Value Curves of Sparse NCPs on Third-order ERP Tensor With Fixed Time Limit of 120s.}
\label{Fig:ERP3rdObj}
\end{figure*}

We also recorded the objective function values of all methods within the first 120 seconds. The results of these methods with $\beta_n=0,10,50,100$ are displayed in \textbf{Fig. \ref{Fig:ERP3rdObj}}. \textbf{Fig. \ref{Fig:ERP3rdObj}} shows that MU still converges very slowly and ALS is not stable. When sparse regularization is added, HALS has very large objective function value. ANLS-AS and ANLS-BPP converge fast at the beginning, but have slightly higher objective function values with $\beta_n > 0$ due to the squared $l_1$-norm regularization item. APG has excellent performances in both speed and convergence for this third-order ERP tensor.

\begin{figure*}[!t]
\centering
\includegraphics[width=\linewidth]{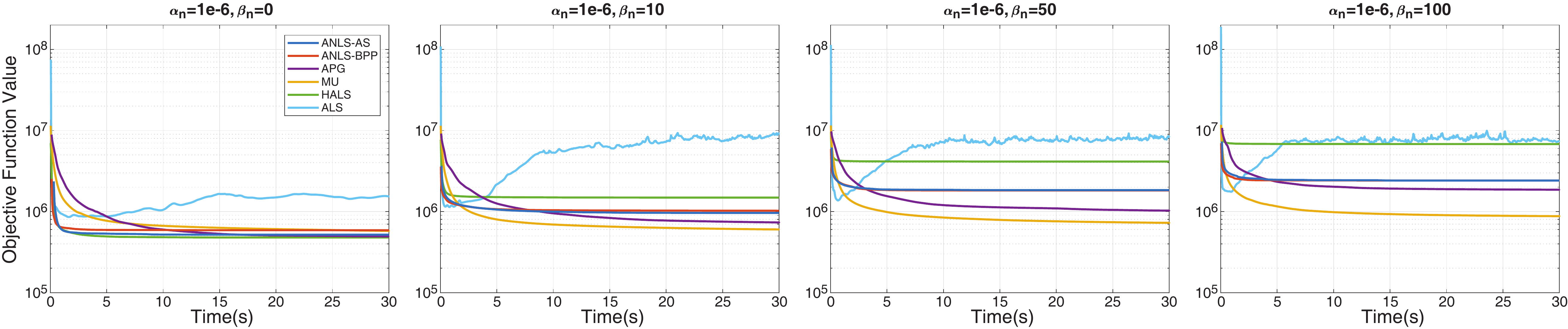}
\caption{The Objective Function Value Curves of Sparse NCPs on Fourth-order ERP Tensor With Fixed Time Limit of 30s.}
\label{Fig:ERP4thObj}
\end{figure*}

\begin{figure*}[!t]
\centering
\includegraphics[width=\linewidth]{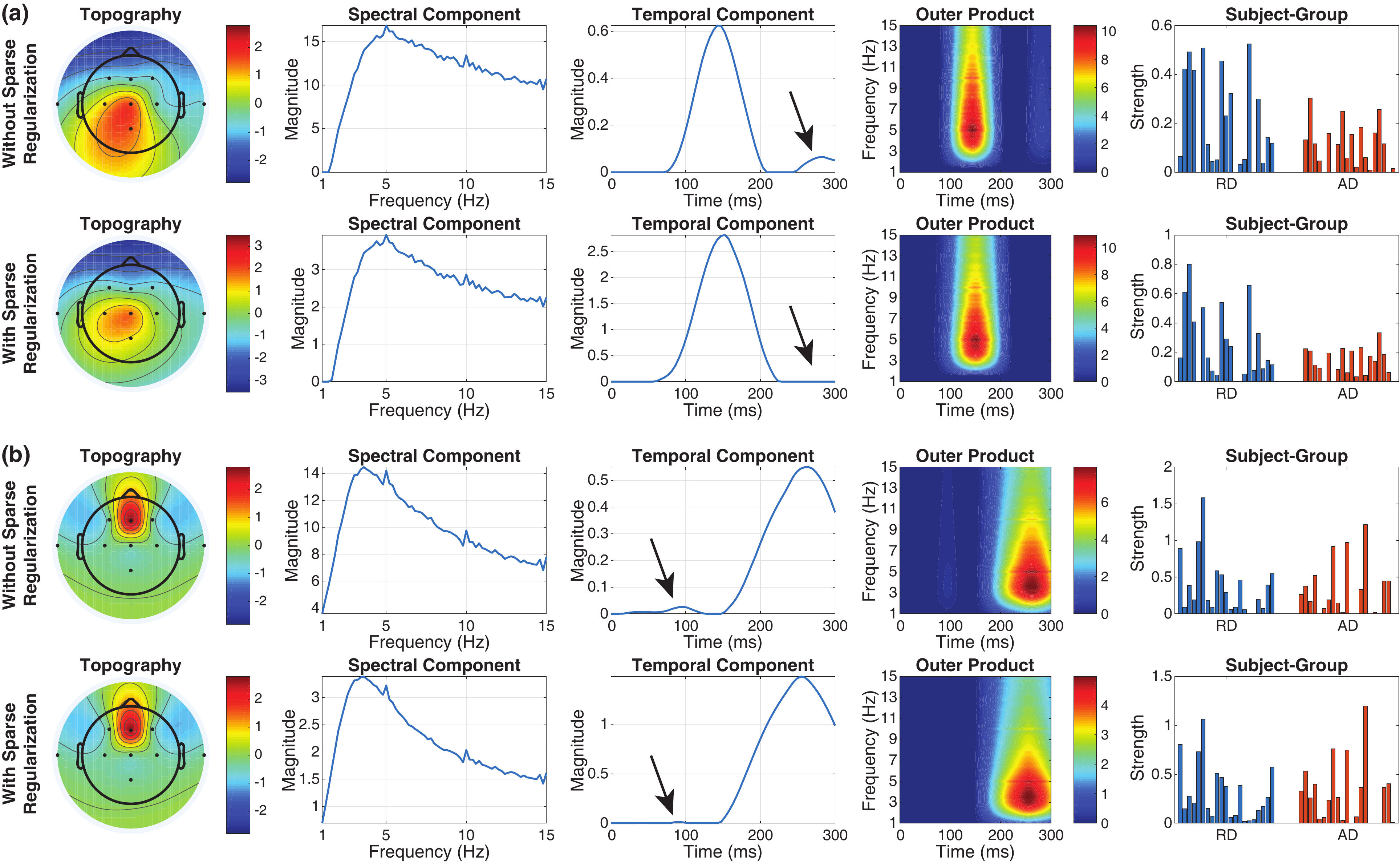}
\caption{Selected components from the fourth-order ERP tensor without and with sparse regularization. The top components in (a) and (b) were extracted using original APG method. The bottom components in (a) and (b) were extracted from sparsity regularized APG with $\beta_n=10$.}
\label{Fig:ERP4thComponents}
\end{figure*}

\subsection{Fourth-Order ERP Data}
\label{Section:ERP4th}
In the fourth experiment, we applied all the sparse NCP algorithms to the original fourth-order ERP tensor introduced in subsection \ref{Section:ERP3rd}. For this experiment, the maximum running time was set by 120 seconds. Other settings are the same as those in the third-order case. We recorded the values of objective function, relative error, running time, iteration number, the sparsity level and nonzero component number of the temporal factor matrix. The average values after 30 times of run are recorded in \textbf{Table \ref{Table:ERP3rdAnd4th}}. We also recorded the objective function values of all methods with $\beta_n=0,10,50,100$ within the first 30 seconds as shown in \textbf{Fig. \ref{Fig:ERP4thObj}}.

For ANLS-AS, ANLS-BPP, MU, HALS and ALS, the results of the fourth-order ERP tensor in \textbf{Table \ref{Table:ERP3rdAnd4th}} and \textbf{Fig. \ref{Fig:ERP4thObj}} reveal similar properties as those in the third-order case. For the fourth-order case, MU is not very effective to impose sparsity on the factor matrices, whose sparsity are very close to the results without sparse regularization. Hence MU has smaller relative error and lower objective function value as shown in \textbf{Fig. \ref{Fig:ERP4thObj}}. However, APG minimizes the objective function slowly even than MU for the fourth-order ERP tensor, which is quite different from the situation in the third-order case.

Despite the slow decrease of the objective function, APG can guarantee to converge and has exhibited effectiveness to impose sparsity. Two groups of components extracted by APG method without and with sparse regularization are shown in \textbf{Fig. \ref{Fig:ERP4thComponents}}. It is clear to see from the temporal components that, with proper sparse regularization parameter $\beta_n$ ($\beta_n=10$ in this case), some weak and redundant elements are suppressed.

\section{Discussion}
From the results and analyses of the sparse NCP experiments, we have the following findings.

MU is a common method for NMF, although it shows slow convergence. It can also be extended to NCP due to the flexibility to handle regularization. However, MU still convergences very slowly in NCP compared with other methods, which sometimes is not very sensitive to sparse regularization.

ALS method is very simple to implement, but it is not stable for NCP with sparse regularization. By ALS, the nonnegative constraint is just obtained by projecting all negative elements of factor matrices to zeros in the subproblems. Hence the solution by projecting is not an accurate solution to subproblem. Therefore, ALS method can not guarantee the convergence of NCP.

HALS is a very efficient method for NCP. Nevertheless, the required normalization procedures will complicate the optimization problem. In our experiments, we find that HALS does not perform well to impose sparsity. High objective function value and relative error occur with large sparse regularization parameters due to the normalization procedures.

APG has proved to be a convergent and stable method for tensor decomposition due to the proximal gradient, which exhibits high efficiency for third-order tensor. Meanwhile, APG can also handle sparse regularization flexibly, which yields closed-form solution. In spite of the convergence property, APG turns out very slowly for the fourth-order tensor case. In future, further studies are needed to accelerate APG for higher-order (order$\geqslant$4) nonnegative tensor decomposition.

ANLS framework converges fast and stably for sparse NCP benefiting from the advantages of many NNLS optimization methods, such as Active Set (AS) and Block Principal Pivoting (BPP). In addition, ANLS exhibits excellent performance to impose sparsity by tuning the sparse regularization parameter $\beta_n$. However, in order to be compatible with the ANLS framework, squared $l_1$-norm has to be employed as the sparse regularization item. Therefore, the method to impose sparsity in ANLS is slightly different from those in other sparse NCP methods. It is very interesting to find the way to solve sparse NCP with non-squared $l_1$-norm by NNLS, which will further minimize the objective function value as we believe.

At last, we want to mention the convergence properties of above optimization methods. The Frobenius norm regularization in sparse NCP \eqref{Eq:SparseNCP} can prevent rank deficiency of the matrix $\bm{B}^{(n)}$ in subproblem \eqref{Eq:SparseNCPSub}, which will improve the convergence of sparse NCP to a globally optimal solution \cite{Lim2009}. ANLS-BPP and ALS method require the $\bm{B}^{(n)}$ to be of full rank \cite{Kim2014}, while the methods of ANLS-AS, APG, HALS and MU don't require the full rank condition. Both ANLS framework and APG method have very good convergence properties, which have been explained in \cite{Kim2014} and \cite{Xu2013} respectively. The subproblem of HALS method has closed form solution \cite{Kim2014}, but the normalization procedures might spoil the convergence property of HALS \cite{Lin2007}. In addition, the convergence of MU and ALS can't be guaranteed. We don't make deep analysis of the convergence properties of these optimization methods in this paper. More information about the convergence properties of the general block coordinate descent (BCD) method can be found in \cite{Bertsekas2016}.

\section{Conclusion}
In this paper, we investigated CANDECOMP/PARAFAC tensor decomposition with both nonnegative constraint and sparse regularization (sparse NCP). The methods of MU, ALS, HALS, APG and ANLS in block coordinate descent framework were deeply analyzed to solve sparse NCP. We compared all these methods by experiments on synthetic and real tensor data, both of which contain third-order and fourth-order cases. We find that APG and ANLS methods are stably convergent and highly effective to impose sparsity for the tensor decomposition compared with other methods. Meanwhile, ANLS framework is very efficient to process higher-order (order$\geqslant$4) tensor data. The proposed accelerated method to compute the objective function and relative error had further improved the efficiency of the sparse NCP.



%




\ifCLASSOPTIONcaptionsoff
  \newpage
\fi



\bibliographystyle{IEEEtran}
\bibliography{IEEEabrv,Constrained_Tensor_with_DOI}
%
%
%

%




\end{document}